\PassOptionsToPackage{numbers,sort&compress}{natbib}

\documentclass[10pt,twocolumn,letterpaper]{article}

\usepackage[pagenumbers]{cvpr} 

\usepackage{float}
\usepackage{siunitx}
\usepackage{xspace}
\usepackage{adjustbox}
\usepackage{multirow}
\usepackage{color, colortbl}
\usepackage{makecell}

\usepackage[accsupp]{axessibility}

\newcommand{\methodname}{Vista4D\xspace}
\newcommand{\website}{https://eyeline-labs.github.io/Vista4D}
\newcommand{\projectpage}{\href{\website}{project page}}

\newcommand{\Paragraphnoskip}[1]{\noindent \textbf{#1.}}
\newcommand{\Paragraph}[1]{\smallskip\noindent \textbf{#1.}}

\newcommand{\NN}{\mathcal{N}}
\newcommand{\LL}{\mathcal{L}}

\renewcommand{\vec}{\mathbf}
\newcommand{\src}{\mathrm{src}}
\newcommand{\tgt}{\mathrm{tgt}}
\newcommand{\gen}{\mathrm{gen}}

\newcommand{\stc}{\mathrm{stc}}
\newcommand{\img}{\mathrm{img}}

\newcommand{\srctgt}{\mathrm{src \to tgt}}

\newcommand{\tgtsrc}{\mathrm{tgt \to src}}

\newcommand{\tss}{\textsuperscript}

\newcommand{\up}{$\Uparrow$}
\newcommand{\down}{$\Downarrow$}

\usepackage{microtype}

\definecolor{cvprblue}{rgb}{0.21,0.49,0.74}
\usepackage[pagebackref,breaklinks,colorlinks,allcolors=cvprblue]{hyperref}
\setcitestyle{numbers,sort&compress}

\def\paperID{15091} 
\def\confName{CVPR}
\def\confYear{2026}

\title{\methodname: Video Reshooting with 4D Point Clouds\vspace{-0.5em}}

\author{
    Kuan Heng Lin\tss{1,3$\dagger$}\quad
    Zhizheng Liu\tss{1,4$\dagger$}\quad
    Pablo Salamanca\tss{1,2}\quad
    Yash Kant\tss{1,2}\\
    Ryan Burgert\tss{1,2,5$\dagger$}\quad
    Yuancheng Xu\tss{1,2}\quad
    Koichi Namekata\tss{1,2,6$\dagger$}\quad
    Yiwei Zhao\tss{2}\\
    Bolei Zhou\tss{4}\quad
    Micah Goldblum\tss{3}\quad
    Paul Debevec\tss{1,2}\quad
    Ning Yu\tss{1,2}\vspace*{0.5em}\\
    {
        \normalsize
        \tss{1}Eyeline Labs\quad
        \tss{2}Netflix\quad
        \tss{3}Columbia University\quad
        \tss{4}UCLA\quad
        \tss{5}Stony Brook University\quad
        \tss{6}University of Oxford\vspace*{0.5em}
    }
}

\begin{document}

\twocolumn[{
    \renewcommand\twocolumn[1][]{#1}
    \maketitle
    \vspace{-1.5em}
    \centerline{\large\textbf{\texttt{\textcolor[RGB]{0,127,255}{\href{https://eyeline-labs.github.io/Vista4D}{eyeline-labs.github.io/Vista4D}}}}}
    \vspace{1.0em}
    \begin{center}
        \centering
        \captionsetup{type=figure}
        \makebox[\linewidth][c]{\includegraphics[width=7.25in]{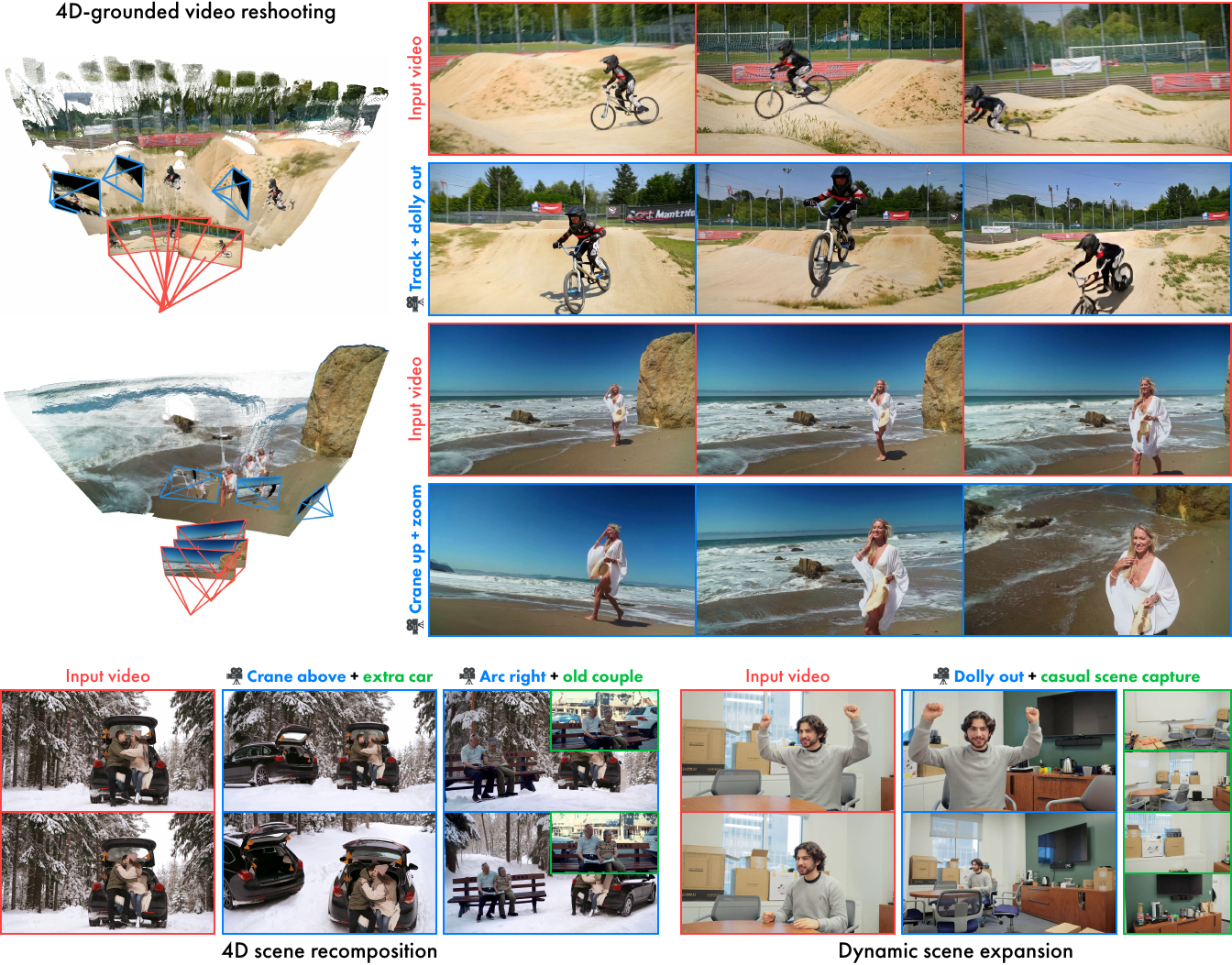}}
        \captionof{figure}{\Paragraphnoskip{4D-grounded video reshooting} Given an input video, \methodname re-synthesizes the scene with the same dynamics from different camera trajectories and viewpoints by grounding the input video and target cameras in a 4D point cloud. \methodname is robust to point cloud artifacts and generalizes to real-world applications such as 4D scene recomposition and dynamic scene expansion.}
        \vspace{-0.5em}
        \label{fig:teaser}
    \end{center}
    \vspace{1em}
    \noindent\makebox[0.5\columnwidth][l]{\rule{0.5\columnwidth}{0.4pt}}\\
    \vspace{-0.5em}
    \footnotesize{\tss{$\dagger$}Work done during an internship at Eyeline Labs.}
    \vspace{1em}
}]

\begin{abstract}

We present \methodname, a robust and flexible video reshooting framework that grounds the input video and target cameras in a 4D point cloud. Specifically, given an input video, our method re-synthesizes the scene with the same dynamics from a different camera trajectory and viewpoint. Existing video reshooting methods often struggle with depth estimation artifacts of real-world dynamic videos, while also failing to preserve content appearance and failing to maintain precise camera control for challenging new trajectories. We build a 4D-grounded point cloud representation with static pixel segmentation and 4D reconstruction to explicitly preserve seen content and provide rich camera signals, and we train with reconstructed multiview dynamic data for robustness against point cloud artifacts during real-world inference. Our results demonstrate improved 4D consistency, camera control, and visual quality compared to state-of-the-art baselines under a variety of videos and camera paths. Moreover, our method generalizes to real-world applications such as dynamic scene expansion and 4D scene recomposition.

\end{abstract}
\vspace*{-1em}\section{Introduction}
\label{sec:intro}

The camera is the visual portal to the filmmaker's world, guiding the audience's gaze as the story unfolds and constructing the narrative's visual language. While traditional visual effects can dramatically transform a raw film set into an immersive spectacle, the ability to manipulate the camera during post-production introduces another dimension of control over visual storytelling.

To this end, we synthesize or `render' the dynamic scene specified by an input source video from novel camera trajectories and viewpoints, which we call \emph{video reshooting}. Importantly, we must achieve faithful reconstruction of seen content in the source video and photorealistically plausible generation of unseen content, all while maintaining precise, user-definable camera control.

We will employ video diffusion models since they are powerful priors for generating dynamic content which is geometrically and temporally coherent \cite{brooks2024sora, wan2025, wiedemer2025videozeroshot, kong2025hunyuanvideosystematicframeworklarge, nvidia2025worldsimulationvideofoundation, genmo2024mochi}.  We will further combine the diffusion models with 4D reconstruction which lifts the monocular source video into a 4D point cloud, providing spatiotemporal grounding for reconstruction and a rich signal for camera control. We present \emph{\methodname}, a video reshooting framework that grounds the source video and target cameras in a 4D point cloud with temporally-persistent static pixels, while leveraging the generative priors of video diffusion models.

Existing works for video reshooting \cite{yu2025trajcraft, ren2025gen3c, hu2025ex4d} condition video diffusion models on per-frame depth-lifted point clouds rendered in the target cameras. However, they often struggle with geometry artifacts and/or temporal flickering due to imprecise 4D reconstruction of real-world dynamic videos as they are often trained on point cloud renders from precise depth maps. Moreover, they also struggle with accurate camera control and content preservation with challenging target camera trajectories and viewpoints.

\methodname introduces the following key designs that not only show state-of-the-art visual quality and robustness to a wide variety of source videos and target cameras but also extend our method with capabilities beyond vanilla video reshooting. First, we build a 4D-grounded point cloud representation where static pixels are visible from any frame via segmentation and 4D reconstruction, as opposed to the per-frame 3D point cloud of baselines. Conditioning with static pixel temporal persistence establishes both explicit preservation of seen content and provides rich camera signals even when the target cameras have little per-frame overlap with the source video. Second, we augment model training with dynamic, 4D-reconstructed multiview video pairs that contain depth estimation artifacts from non-frontal views. Thus, \methodname is significantly more robust to the quality of real-world point cloud renders while allowing us to additionally condition on the source video to utilize video model priors for geometric coherence. This further enables us to manipulate the 4D point cloud during inference for real-world applications beyond video reshooting.

Our contributions are as follows:
\begin{itemize}
    \item We present \methodname, a video reshooting framework that maintains geometric and physical plausibility with real-world inference, while explicitly preserving seen content by grounding generation in a 4D point cloud.
    \item Through extensive quantitative and qualitative comparisons, including a user study, we validate the improved content preservation, camera controllability, and visual quality of \methodname over state-of-the-art baselines for a wide variety of videos and cameras.
    \item We show that our training extends \methodname with capabilities that generalize to real-world applications such as dynamic scene expansion, 4D scene recomposition, and long video inference with memory.
\end{itemize}

\vspace*{-0.5em}\section{Related work}\vspace*{-0.5em}
\label{sec:related}

\begin{figure*}  
  \centering
  \includegraphics[width=\textwidth]{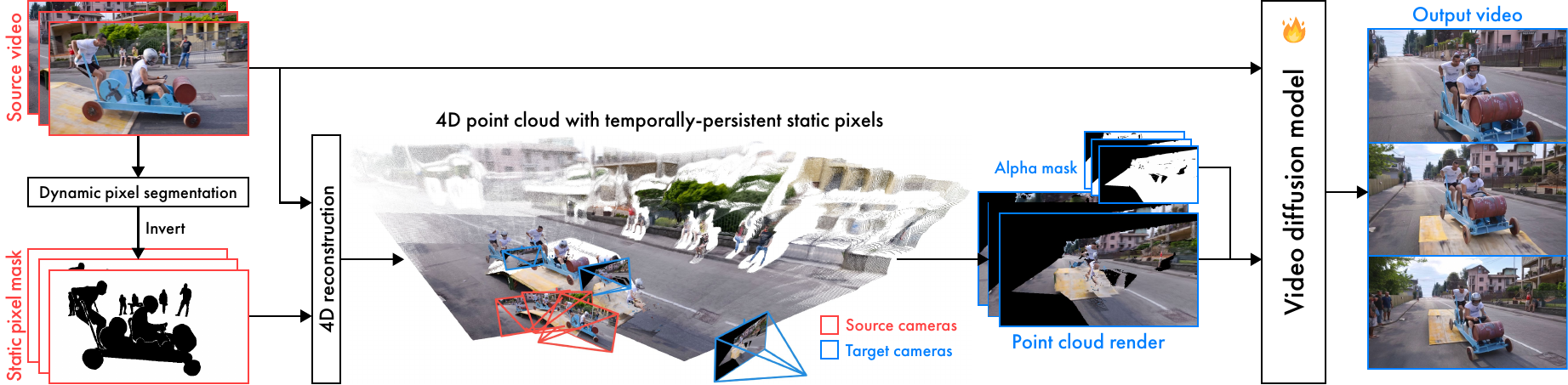}
  \caption{\Paragraphnoskip{Overview of \methodname} Given an input source video, we build a 4D point cloud where static pixels are temporally persistent via segmentation and 4D reconstruction. We then render the point cloud in the target cameras which users define. Lastly, the source video and point cloud render \& alpha mask are jointly processed by the finetuned video diffusion model to generate a video of the same dynamic scene in the target cameras. We provide model architecture details in  Supplementary \ref{supp:architecture}.}
  \label{fig:pipeline}
\end{figure*}

\begin{figure}
  \centering
  \includegraphics[width=\columnwidth]{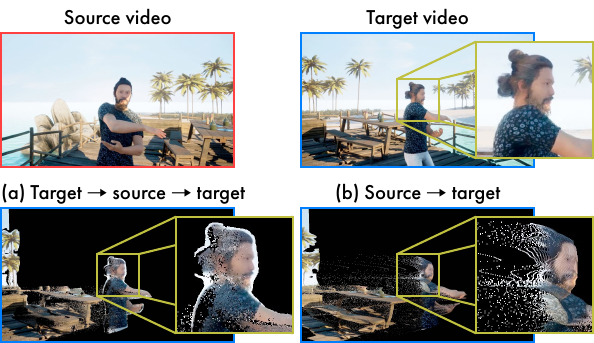}
  \caption{\Paragraphnoskip{Multiview 4D reconstruction artifacts} (a) Double reprojection \cite{yu2025trajcraft} first renders the target video point cloud in the source camera, then rerendering it in the target camera to create occluded regions for paired training, thus viewing the target video depth map from its frontal, artifact-free view. (b) In contrast, rendering the source video point cloud from the target camera with dynamic multiview data exposes non-frontal-view artifacts that better match real-world inference. The above source-target video pair is from MultiCamVideo \cite{bai2025recammaster} with 4D reconstruction by STream3R \cite{lan2025stream3r}.}
  \label{fig:multiview}
\end{figure}

\Paragraph{Video reshooting with \emph{explicit priors}}
For video reshooting, and more broadly novel view synthesis of static scenes, 3D/4D point clouds provide an explicit and rich spatial prior. To this end, video reshooting methods with \emph{explicit priors} \cite{ren2025gen3c, yu2025trajcraft, hu2025ex4d, jeong2025reangle, you2025nvssolver, qian2025wristworld} use video depth estimators \cite{chen2025videodepthanything, hu2025depthcraft, xu2025geocraft} to render per-frame camera-space point clouds as conditioning signals for video diffusion models. Depth estimation priors have also been widely used for static scene novel view synthesis (NVS) \cite{kant2023invs, muller2024multidiff} and video motion control \cite{xiao2025trajattn, ren2025gen3c}. However, many of these methods often train on precise depth maps which inhibits generalization to imperfect real-world depth estimation, and their per-frame point cloud conditioning can struggle to preserve seen content and maintain accurate camera control with challenging camera trajectories.

\Paragraph{Video reshooting with \emph{implicit priors}}
Alternatively, video shooting methods can also use \emph{implicit priors} for camera control such as camera embeddings \cite{bai2025recammaster, vanhoorick2024gcd} or video references \cite{luo2025camclonemaster} by finetuning video diffusion models on time-synchronized synthetic multiview data. Image- and camera-conditioned diffusion models have also been used for static scene NVS~\cite{zhou2025seva, kant2024spad, liu2023zero123, shi2023mvdream,voleti2024sv3d} and camera-controlled video generation \cite{he2025cameractrlii,bahmani2024vd3d,xie2024sv4d}. However, due to the inherent depth scale ambiguity of monocular videos, camera control from implicit-prior methods is often imprecise and unable to be explicitly `previewed' unlike point clouds.

\Paragraph{4D reconstruction}
To provide explicit geometric priors for video reshooting, we lift the input video into a world-space point cloud with 4D reconstruction. Traditional structure from motion~\cite{schonberger2016colmap,wang2024vggsfm, pan2024glomap} rely on multiview geometry constraints but are not robust to dynamic scenes. With the strong performance of learning-based video depth estimation models~\cite{chen2025videodepthanything, hu2025depthcraft, xu2025geocraft, chou2025flashdepth,wang2025moge, piccinelli2025unidepthv2}, recent works \cite{li2025megasam, yao2025uni4d, huang2025vipe} combine these depth priors and camera optimization with SLAM~\cite{teed2021droid} to obtain robust and coherent dynamic scene reconstruction. Followed by recent success in end-to-end 3D reconstruction methods~\cite{wang2024dust3r, wang2025vggt, keetha2025mapanything}, end-to-end 4D reconstruction models~\cite{wang2025pi3, lan2025stream3r, sun2024monst3r, zhuo2025stream4dvgt, jiang2025geo4d} have also emerged as more efficient alternatives. Some recent methods also predict 4D Gaussians from monocular videos~\cite{lei2025mosca, wang2025shapeofmotion, wang2025gflow}, enabling novel view synthesis at small viewpoint deviations from the input videos.

\section{4D-grounded video reshooting}
\label{sec:method}

Given an input source video $\mathbf{X}^\src$, we first build a 4D point cloud via 4D reconstruction with temporally-persistent static pixels defined by static pixel masks from segmentation. We then render the point cloud from the target cameras and jointly condition the finetuned video diffusion model on the source video and point cloud render, producing the output video. Section \ref{sec:method:point_cloud} builds the temporally-persistent point cloud; Section \ref{sec:method:multiview} explains the importance of training with noisily-reconstructed multiview data; Section \ref{sec:method:source} discusses joint conditioning of source videos and point cloud renders; and Section \ref{sec:method:details} describes data and training details. Our method is illustrated in Figure \ref{fig:pipeline}.

\subsection{Building a temporally-persistent 4D point cloud}
\label{sec:method:point_cloud}

To explicitly preserve seen content in the source video and provide more accurate camera control especially when target cameras have little per-frame overlap with the source video, we build a temporally-persistent 4D point cloud. We first use 4D reconstruction \cite{wang2025pi3, lan2025stream3r} to obtain depths $\mathbf{D}^\src$, camera extrinsics $\mathbf{T}^\src$, and camera intrinsics $\mathbf{K}^\src$, and we use segmentation \cite{ravi2024sam2, ren2024groundingdino, ren2024groundedsam} to obtain a static pixel mask $\mathbf{M}^\stc$. We then lift the source video into a world-space per-frame 3D point cloud
\begin{equation}\label{eqn:img_to_wld}
    \vec{P} = \Omega\left( \Phi^{-1}\left( [\vec{X}^\src, \vec{D}^\src], \vec{K}^\src \right), \vec{T}^\src \right) ,
\end{equation}
where $\Phi^{-1}$ and $\Omega$ are the inverse perspective projection and world-space transformation. Since the per-frame point cloud $\vec{P}$ is grounded in world space, we use $\mathbf{M}^\stc$ to make static pixels persistent across all frames to incorporate explicit 4D context in our point cloud rendering, obtaining the temporally-persistent point cloud $\overline{\mathbf{P}}$. Then, we render $\overline{\mathbf{P}}$ from the target cameras, obtaining the point cloud render $\vec{X}^\srctgt$ and its alpha mask $\vec{M}^\srctgt$ as temporally persistent, 4D-grounded priors for the video diffusion model.

\subsection{Training with noisy multiview data}
\label{sec:method:multiview}

So far, generating our 4D point cloud requires source-target video pairs: The source video builds the temporally-persistent point cloud, and the target video defines the target cameras. Because 4D reconstruction methods are imperfect, the point cloud render during inference often contain \emph{geometric artifacts} when the target cameras deviate far from the frontal view of the lifted point cloud. This is especially true for dynamic pixels where depth estimators cannot leverage multiview geometry constraints from moving cameras. Existing methods \cite{yu2025trajcraft, ren2025gen3c, hu2025ex4d} instead train with artifact-free point clouds, which essentially simplify video reshooting to inpainting. For example, as illustrated in Figure \ref{fig:multiview} (a), TrajectoryCrafter \cite{yu2025trajcraft} applies double-reprojection to monocular videos to obtain paired data of point cloud renders and target videos, which always views the depth maps from their frontal, artifact-free view. In contrast, we train with multiview dynamic-scene videos with 4D-reconstructed depths and cameras, which results in spatially mismatching point clouds artifacts compared to the target video as shown in Figure \ref{fig:multiview} (b). Thus, our method moves beyond inpainting and instead corrects imperfect point cloud geometry.

As real-world multiview video datasets with dynamic scenes are rare and small in scale, we use synthetic multiview dynamic videos to train our model as in \cite{bai2025recammaster}. Moreover, to ensure the model is generalizable to real-world video inputs while being robust to noisy 4D reconstruction, we train with a mix of multiview synthetic and real-world monocular data. For monocular data, following TrajectoryCrafter \cite{yu2025trajcraft}, we first render the point cloud of the target video from heuristic-generated source cameras to produce $\vec{X}^\tgtsrc$. Then, we render $\vec{X}^\tgtsrc$ back to the original target cameras to produce the double-reprojected point cloud render.

\subsection{Conditioning on source videos and point clouds}
\label{sec:method:source}

Point cloud artifacts during real-world inference obfuscate not only geometry but also appearance information from the source video. Thus, while some existing methods only condition on point cloud renders \cite{ren2025gen3c, hu2025ex4d}, we also condition on source videos to utilize video diffusion model priors for transferring geometric and appearance information like implicit-prior methods do \cite{bai2025recammaster, luo2025camclonemaster}. Unlike TrajectoryCrafter's cross-attention injection of source videos \cite{yu2025trajcraft}, we concatenate the patchified latent tokens of the source video and point cloud render with the noisy target latent tokens along the frame dimension. We find that in-context conditioning best preserves source video content and is thus more robust to point cloud artifacts, which we ablate in Supplementary \ref{supp:ablations}.

Thus, given the source video $\mathbf{X}^\src$, point cloud render $\mathbf{X}^\srctgt$ and its alpha mask $\vec{M}^\srctgt$, and target cameras $\vec{C}^\tgt = \left( \vec{K}^\tgt, \vec{T}^\tgt \right)$, we finetune a video diffusion transformer $\boldsymbol{\epsilon}_\theta$ to generate the target video $\vec{X}^\tgt$ with the flow matching objective
\begin{equation}
    \LL = \left\lVert \boldsymbol{\epsilon}_\theta \left( \vec{X}^\tgt_t, \vec{X}^\srctgt, \vec{M}^\srctgt, \vec{X}^\src, \vec{C}^\tgt, t \right) - \vec{V} \right\rVert ,
\end{equation}
where $\vec{V} = \vec{X}^\tgt - \boldsymbol{\epsilon}$ and $\vec{X}^\tgt_t$ is the noisy target video at timestep $t$ by sampled Gaussian $\boldsymbol{\epsilon}$. We inject the target cameras $\vec{C}^\tgt$ as Pl\"{u}cker embeddings \cite{kuang2024collaborative, he2025cameractrl, xu2025virtuallybeing} via zero-initialized linear projections, with an identity-initialized projection after self-attention, inspired by ReCamMaster \cite{bai2025recammaster}. We provide model architecture details in Supplementary \ref{supp:architecture}.

Conditioning the model with both the source video and point cloud render allows the model to learn to propagate geometry and appearance information from the source to the output video. For monocular training videos without a ground-truth $\vec{X}^\src$, we condition the model on $\vec{X}^\tgtsrc$ as an occluded source video with its alpha mask to still learn this propagation.

\begin{table}
    \caption{\Paragraphnoskip{Camera control accuracy and 3D consistency} \methodname consistently shows the most accurate camera control compared to baselines with superior rotation, translation, and intrinsics errors. Our method also significantly outperforms baselines in per-frame 3D consistency with the lowest reprojection error under SuperGlue (RE@SG) \cite{sarlin2020superglue, detone2018superpoint, kant2025pippo}. \textbf{Bold} indicates best results.}
    \label{tab:camera_accuracy}
    \centering
    \begin{adjustbox}{max width=\columnwidth}
        \begin{tabular}{l c c c c}
            \toprule
            Method & \makecell{Translation \\ error \down} & \makecell{Rotation \\ error \down} & \makecell{Intrinsics \\ error \down} & \makecell{RE@SG \\ \down} \\
            \midrule
            ReCamMaster \cite{bai2025recammaster} & 1.574 & 12.79 & 11.16 & 23.66 \\
            CamCloneMaster \cite{luo2025camclonemaster} & 2.132 & 23.77 & 6.422 & 23.38 \\
            TrajectoryCrafter \cite{yu2025trajcraft} & 1.434 & 6.838 & 6.671 & 120.5 \\
            EX-4D \cite{hu2025ex4d} & 1.325 & 5.941 & 5.182 & 13.11 \\
            GEN3C \cite{ren2025gen3c} & 1.309 & 4.751 & 5.085 & 12.99 \\
            \rowcolor{gray!10} \textbf{\methodname (ours)} & \textbf{1.251} & \textbf{4.647} & \textbf{4.927} & \textbf{7.504} \\
            \bottomrule
        \end{tabular}
    \end{adjustbox}
\end{table}

\begin{table}
    \caption{\Paragraphnoskip{Novel-view video synthesis} \methodname shows comparable to superior noel-view video synthesis performance on the \texttt{iphone} dataset \cite{gao2022dycheck}. EPE (endpoint error) measures optical flow error between the generated and ground truth videos and indicates scene motion reconstruction. \textbf{Bold} indicates best results.}
    \label{tab:nvs}
    \centering
    \begin{adjustbox}{max width=\columnwidth}
        \begin{tabular}{l c c c c c c c}
            \toprule
            Method & \makecell{mPSNR \\ \up} & \makecell{mSSIM \\ \up} & \makecell{mLPIPS \\ \down} & \makecell{PSNR \\ \up} & \makecell{SSIM \\ \up} & \makecell{LPIPS \\ \down} & \makecell{EPE \\ \down} \\
            \midrule
            ReCamMaster \cite{bai2025recammaster} & 10.84 & 0.444 & 0.692 & 10.96 & 0.262 & 0.755 & 4.681 \\
            CamCloneMaster \cite{luo2025camclonemaster} & 11.14 & 0.444 & 0.651 & 11.17 & 0.260 & 0.713 & 4.318 \\ 
            TrajectoryCrafter \cite{yu2025trajcraft} & 13.82 & \textbf{0.492} & 0.569 & 13.06 & \textbf{0.320} & 0.656 & 2.375 \\
            EX-4D \cite{hu2025ex4d} & 12.85 & 0.479 & 0.596 & 12.64 & 0.305 & 0.669 & 4.269 \\
            GEN3C \cite{ren2025gen3c} & 12.19 & 0.447 & 0.608 & 12.06 & 0.260 & 0.679 & 3.019 \\
            \rowcolor{gray!10} \textbf{\methodname (ours)} & \textbf{14.09} & 0.480 & \textbf{0.461} & \textbf{14.14} & 0.310 & \textbf{0.514} & \textbf{1.142} \\
            \bottomrule
        \end{tabular}
    \end{adjustbox}
\end{table}

\begin{table*}
    \caption{\Paragraphnoskip{Video fidelity} \methodname consistently outperform point-cloud-conditioned (explicit-prior) baselines for the video fidelity metrics FID, FVD, CLIP-T, and metrics from VBench \cite{huang2024vbench} and VBench-2.0 \cite{zheng2025vbench2}. Implicit-prior methods (ReCamMaster and CamCloneMaster) outperform our method in some metrics due to their low camera control accuracy (Table \ref{tab:camera_accuracy}) that result in output videos with similar, usually more static, cameras to the input video which produces better FID, FVD, and VBench consistency metrics. \textbf{Bold} indicates best results.}
    \label{tab:fidelity}
    \centering
    \begin{adjustbox}{max width=\textwidth}
        \begin{tabular}{l c c c c c c c c c c}
            \toprule
            \multirow{2}{*}{Method} & \multirow{2}{*}{\makecell{Camera \\ control}} & \multirow{2}{*}{\makecell{FID \\ \down}} & \multirow{2}{*}{\makecell{FVD \\ $\times 10^3$ \down}} & \multirow{2}{*}{\makecell{CLIP-T \\ \up}} & \multicolumn{6}{c}{VBench \cite{huang2024vbench} \& VBench-2.0 \cite{zheng2025vbench2}} \\
            \cmidrule(r){6-11}
            & & & & & \makecell{Aesthetic \\ quality \up} & \makecell{Imaging \\ quality \up} & \makecell{Subject \\ consistency \up} & \makecell{Background \\ consistency \up} & \makecell{Temporal \\ style \up} & \makecell{Human \\ anatomy \up} \\
            \midrule

            ReCamMaster \cite{bai2025recammaster} & Extrinsics & \textbf{94.15} & \textbf{1.203} & 0.319 & 0.552 & 0.701 & \textbf{0.913} & \textbf{0.934} & 0.243 & 0.759 \\
            CamCloneMaster \cite{luo2025camclonemaster} & Ref. video & 101.4 & 1.406 & 0.321 & 0.560 & 0.709 & 0.886 & 0.915 & 0.247 & 0.711 \\
            \midrule
            
            TrajectoryCrafter \cite{yu2025trajcraft} & Point cloud & 125.6 & 1.640 & 0.305 & 0.509 & 0.650 & 0.854 & 0.906 & 0.241 & 0.790 \\
            EX-4D \cite{hu2025ex4d} & Point cloud & 124.6 & 1.481 & 0.296 & 0.480 & 0.660 & 0.849 & 0.894 & 0.226 & 0.687 \\
            GEN3C \cite{ren2025gen3c} & Point cloud & 113.5 & 1.441 & 0.318 & 0.519 & 0.660 & 0.857 & 0.913 & 0.245 & 0.775 \\
            \rowcolor{gray!10} \textbf{\methodname (ours)} & Point cloud & 105.4 & 1.418 & \textbf{0.326} & \textbf{0.567} & \textbf{0.716} & 0.883 & 0.916 & \textbf{0.253} & \textbf{0.857} \\
            \bottomrule
        \end{tabular}
    \end{adjustbox}
\end{table*}

\begin{table}
    \caption{\Paragraphnoskip{User study} Participants consistently perfer \methodname over baselines on source video content preservation, camera control accuracy, and overall video fidelity. \textbf{Bold} indicates best results.}
    \label{tab:user_study}
    \centering
    \begin{adjustbox}{max width=\columnwidth}
        \begin{tabular}{l c c c}
            \toprule
            Method & \makecell{Source \\ preservation \up} & \makecell{Camera \\ accuracy \up} & \makecell{Overall \\ fidelity \up} \\
            \midrule
            ReCamMaster \cite{bai2025recammaster} & 9.921\% & 1.905\% & 4.365\% \\
            CamCloneMaster \cite{luo2025camclonemaster} & 15.63\% & 6.429\% & 11.03\% \\
            TrajectoryCrafter \cite{yu2025trajcraft} & 0.952\% & 5.952\% & 0.476\% \\
            EX-4D \cite{hu2025ex4d} & 1.587\% & 6.508\% & 0.794\% \\
            GEN3C \cite{ren2025gen3c} & 4.841\% & 11.03\% & 5.952\% \\
            \rowcolor{gray!10} \textbf{\methodname (ours)} & \textbf{67.06\%} & \textbf{68.17\%} & \textbf{77.38\%} \\
            \bottomrule
        \end{tabular}
    \end{adjustbox}
\end{table}

\subsection{Training details and datasets}
\label{sec:method:details}

For the base video generation model, we build off of \texttt{Wan2.1-T2V-14B} \cite{wan2025}, a pretrained text-to-video flow matching \cite{lipman2023flowmatching} video diffusion transformer \cite{peebles2023dit}. We finetune the model at a resolution of $672 \times 384$ for \num[group-separator={,}]{30000} steps, then at $1280 \times 720$ for \num[group-separator={,}]{300} steps, both with $49$-frame videos, a global batch size of $8$, and the AdamW optimizer with a learning rate of \num{1e-5}. We train the patchify layers for $\vec{X}^\src$ and $\vec{X}^\srctgt$, self-attention layers, camera encoders, and projectors, while freezing all other parameters.

\Paragraph{Datasets} For multiview time-synchronized videos, we adopt the synthetic MultiCamVideo dataset from ReCamMaster \cite{bai2025recammaster}, and we run 4D reconstruction across all views with STream3R \cite{lan2025stream3r}. For real-world monocular videos, we adopt a 60K subset from \texttt{OpenVidHD-0.4M} \cite{nan2025openvid} and run 4D reconstruction with $\pi^3$ \cite{wang2025pi3}. For segmenting static pixels, inspired by Uni4D \cite{yao2025uni4d}, we obtain semantic classes with RAM \cite{zhang2023ram}, filter for dynamic subjects/nouns with \texttt{Llama-3.1-8B-Instruct} \cite{llama2024llama3}, and segment per-frame dynamic pixels with Grounded SAM 2 \cite{ravi2024sam2, ren2024groundingdino, ren2024groundedsam} and invert the resulting masks.

\begin{figure*}
    \centering
    \includegraphics[width=\textwidth]{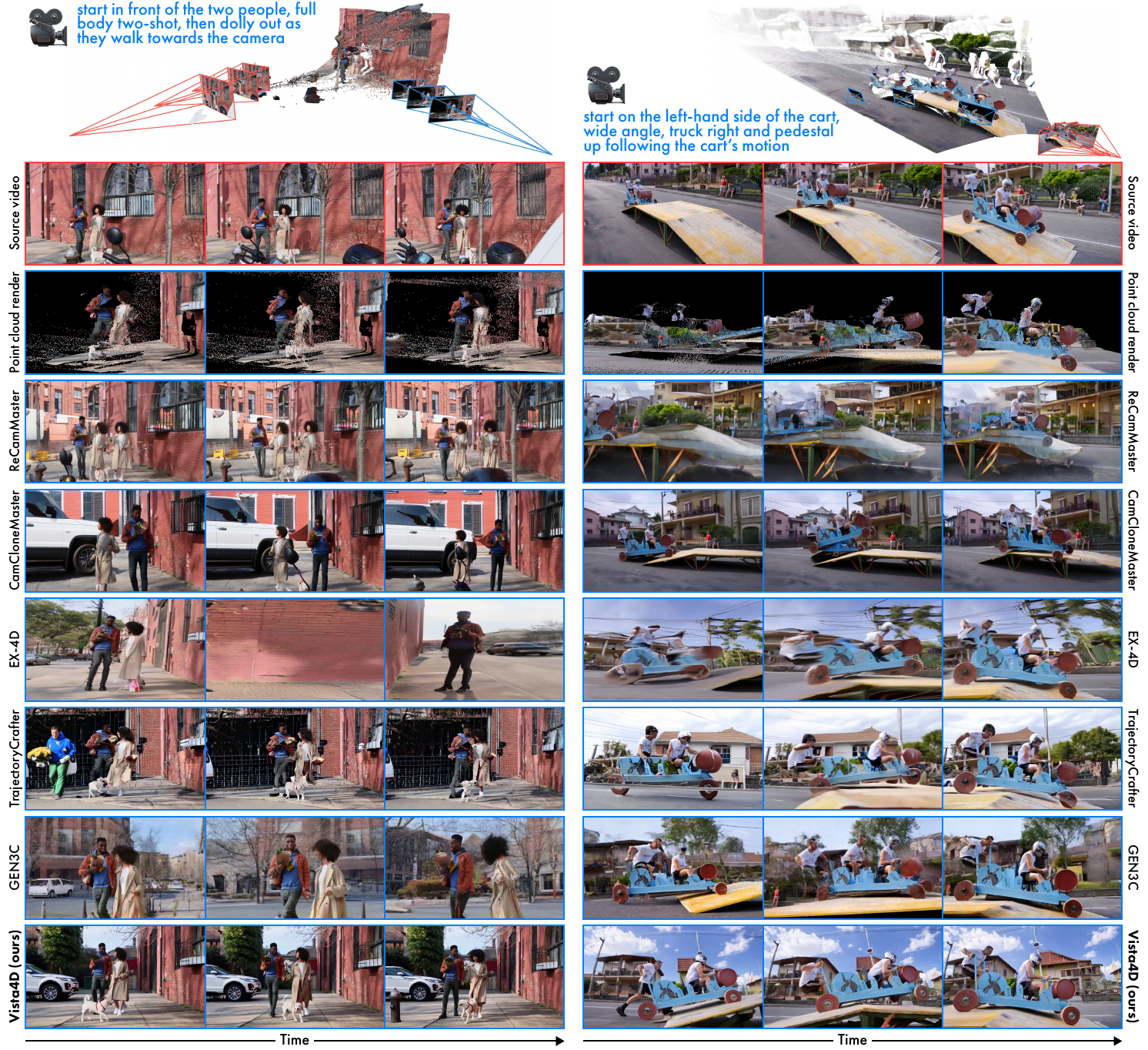}
    \caption{\Paragraphnoskip{Qualitative comparison on real-life monocular videos} We show two video reshooting examples of \methodname compared to our baselines, TrajectoryCrafter \cite{yu2025trajcraft}, GEN3C \cite{ren2025gen3c}, EX-4D \cite{hu2025ex4d}, ReCamMaster \cite{bai2025recammaster}, and CamCloneMaster \cite{luo2025camclonemaster}.}
    \label{fig:qual_comp}
\end{figure*}

\section{Experiments}
\label{sec:results}

\Paragraph{Baselines} We compare \methodname\ to state-of-the-art video reshooting methods. For explicit-prior methods, TrajectoryCrafter \cite{yu2025trajcraft} introduces the double-reprojection technique to generate training pairs from monocular dynamic videos, EX-4D \cite{hu2025ex4d} proposes the Depth Watertight Mesh during inference to train on tracking-based inpainting, and GEN3C \cite{ren2025gen3c} builds a 3D cache with pooling-based fusion for sparse-view novel-view synthesis. For implicit-prior methods, ReCamMaster \cite{bai2025recammaster} constructs a synthetic multiview time-synchronized video dataset to train a camera-conditioned model, and CamCloneMaster \cite{luo2025camclonemaster} replicates camera trajectories from reference videos. We use our $672 \times 384$ checkpoint for all quantitative evaluations and the user study.

\Paragraph{Evaluation dataset}
For quantitative evaluation, we create an evaluation dataset of high quality, diverse $110$ video-camera pairs: We select $51$ videos from DAVIS \cite{perazzi2016davis} and the royalty-free stock video website Pexels \cite{pexels2025}. Then, we run 4D reconstruction with $\pi^3$ \cite{wang2025pi3} and segmentation with Grounded SAM 2 \cite{ravi2024sam2, ren2024groundingdino, ren2024groundedsam}, and we design two to three camera trajectories for each video with our camera design UI, which we show examples of in Supplementary \ref{supp:eval_user}.

\subsection{Quantitative comparisons}
\label{sec:results:quant}

We quantitatively compare \methodname to baselines and show our method's superiority on three dimensions: Camera control and 3D consistency, novel-view video synthesis, and video fidelity. We include details of each quantitative evaluation metric in Supplementary \ref{supp:quant_metrics}.

\Paragraph{Camera control accuracy and 3D consistency}
We compare camera control accuracy and 3D consistency \methodname to baselines in Table \ref{tab:camera_accuracy} on our $110$ video-camera pair dataset. For camera control accuracy, we measure translation, rotation, and intrinsics error between target cameras from the evaluation dataset and 4D-reconstruction-predicted cameras of generated videos from each method \cite{bai2025recammaster, xu2025virtuallybeing}. For 3D consistency between the source and output videos, following Pippo~\cite{kant2025pippo}, we adopt the per-frame reprojection error of SuperPoint \cite{detone2018superpoint} landmarks under SuperGlue (RE@SG) \cite{sarlin2020superglue}. Our method consistently exhibits more accurate camera control compared to baselines, especially against implicit-prior methods. Moreover, our method significantly outperforms baselines in 3D consistency, showcasing its output geometric plausibility despite noisy real-world 4D reconstruction.

\begin{figure}
    \centering
    \includegraphics[width=\columnwidth]{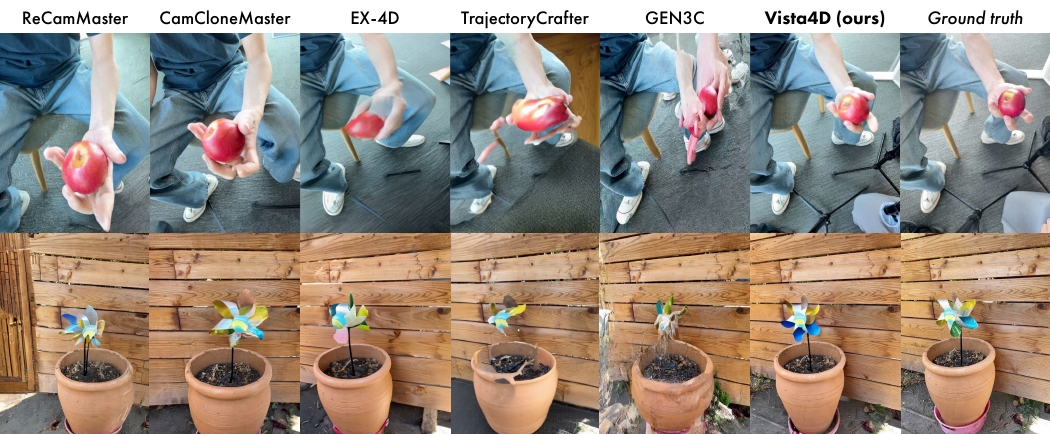}
    \caption{\Paragraphnoskip{Qualitative comparison on novel-view synthesis} We show two samples of \methodname compared our baselines on the \texttt{iphone} dataset \cite{gao2022dycheck}.}
    \label{fig:nvs}
\end{figure}

\Paragraph{Novel-view video synthesis}
We compare novel-view video synthesis quality of \methodname to baselines in Table \ref{tab:nvs} on the real-world time-synchronized multiview dataset, \texttt{iphone} \cite{gao2022dycheck}. We measure masked (indicated by the prefix ``m'') and full PSNR, SSIM, and LPIPS for synthesis quality \cite{gao2022dycheck}, along with optical flow endpoint error (EPE) for motion quality \cite{burgert2025gwtf}. Our method outperforms baselines in PSNR and LPIPS, indicating our superior spatial reconstruction quality. We also significantly outperform baselines for EPE, indicating our method's ability to preserve source video motion. Note that even though we are behind TrajectoryCrafter \cite{yu2025trajcraft} for SSIM, viewing the synthesized videos quickly reveal significant artifacts in the latter's outputs not caught by SSIM, examples of which we show in Figure \ref{fig:nvs}.

\Paragraph{Video fidelity}
We evaluate video fidelity and quality of \methodname to baselines in Table \ref{tab:fidelity} on our $110$ video-camera pair dataset. We use FID \cite{heusel2017fid}, FVD \cite{uterthiner2019fvd}, VBench (aesthetic quality, imaging quality, subject consistency, background consistency, and temporal style) \cite{huang2024vbench}, and VBench-2.0 (human anatomy) \cite{zheng2025vbench2} to evaluate video fidelity, and CLIP-T \cite{radford2021clip} for prompt alignment. Our method consistently outperforms point-cloud-conditioned (explicit-prior) baselines, especially in aesthetic quality, imaging quality, and human anatomy due to our robustness to point cloud artifacts. Implicit-prior methods (ReCamMaster and CamCloneMaster) perform better in FID, FVD, subject consistency, and background consistency because they often fail to follow the target cameras, resulting in relatively little camera change from the source video, which results in seemingly better metrics. Qualitative comparisons in \ref{fig:qual_comp} and Supplementary \ref{supp:qualitative}, along with the user study in Table \ref{tab:user_study}, show our method's clear high video fidelity.

\subsection{Qualitative comparisons}
\label{sec:results:qual}

We qualitatively compare \methodname to baselines in Figure \ref{fig:qual_comp} on two example real-life monocular videos, where we show the point cloud render to illustrate the intended cameras in addition to the written description. Explicit-prior methods (EX-4D, TrajectoryCrafter, and GEN3C) all struggle with point cloud artifacts from target cameras at non-frontal views of the depth maps, resulting in subject and background artifacts (\eg, TrajectoryCrafter, left video; all three methods, right video) or camera control failure (\eg, EX-4D and GEN3C, left video). Implicit-prior methods (ReCamMaster and CamCloneMaster) similarly struggle with precise camera control (ReCamMaster, left video; both methods, right video) and subject artifacts (both methods, left video). In contrast, our method produces high-quality outputs that not only faithfully preserves source video content but also follows the target cameras. We include more comprehensive qualitative results and comparisons in Supplementary \ref{supp:qualitative}.

\Paragraph{User study}
We show the results of our user study in Table \ref{tab:user_study}, where we ask participants to compare our method to baselines on three dimensions: Source video content preservation, camera control accuracy, and overall video fidelity. We randomly select a subset of $30$ video-camera pairs from the $110$-pair evaluation dataset and invite $42$ participants to select their preferred method for each pair and each dimension. Users consistently prefer our method by a wide margin over baselines on all dimensions, especially overall video fidelity due to our method's robustness to point cloud artifacts and challenging camera trajectories and viewpoints. We include details of our user study in Supplementary \ref{supp:eval_user}.

\begin{figure}
    \centering
    \includegraphics[width=\linewidth]{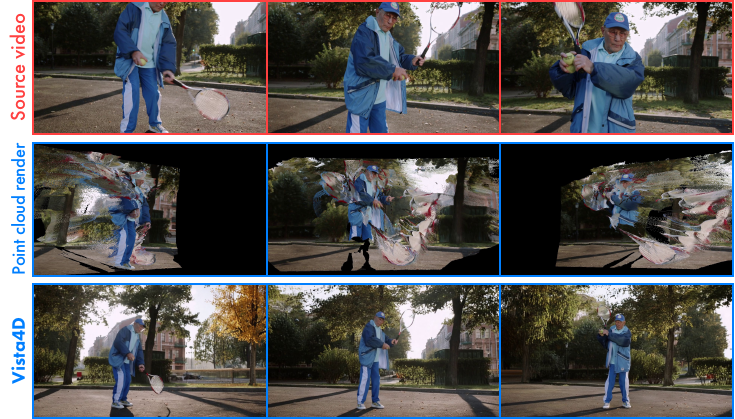}
    \caption{\Paragraphnoskip{Robustness to segmentation failure} We simulate segmentation failure by not segmenting the tennis racket as dynamic. \methodname is generally robust to these point cloud streaks as it utilizes the in-context-conditioned source video to correct the artifacts.}
    \label{fig:seg_fail}
\end{figure}

\Paragraph{Robustness to segmentation failure}
Since \methodname uses Grounded SAM 2 to segment dynamic pixels, segmentation failures can result in point cloud streaking artifacts. However, \methodname is generally robust to them. For example, we simulate segmentation failure in Figure \ref{fig:seg_fail} by deliberately not segmenting the tennis racket, and \methodname corrects the streaking just like it corrects imperfect point cloud geometry, that is, by utilizing the in-context-conditioned source video. Broadly, we observe that streaking artifacts during inference are rare or inconsequential, especially compared to the improved camera control and 4D consistency from static pixel temporal persistence.

\subsection{Ablation study}
\label{sec:results:ablation}

We study the effects of our data and model conditioning on source video content preservation and robustness to imperfect 4D reconstruction. We perform ablations combinations of the following design choices: No depth artifacts (by always doing double reprojection), no source video, cross-attention source video injection, and no temporal persistence. We find that the combination of training with depth artifacts and the (in-context conditioned) source video enables our model's ability to be robust to 4D reconstruction artifacts, particularly both spatial artifacts (imprecise depths from non-frontal views) and temporal artifacts (jittering depths). We also find that removing temporal persistence reduces our model's ability to both preserve static content and maintain accurate camera control when the source video and target cameras have low per-frame overlap. We show examples of both findings in Supplementary \ref{supp:ablations}.

\begin{figure}
    \centering
    \includegraphics[width=\columnwidth]{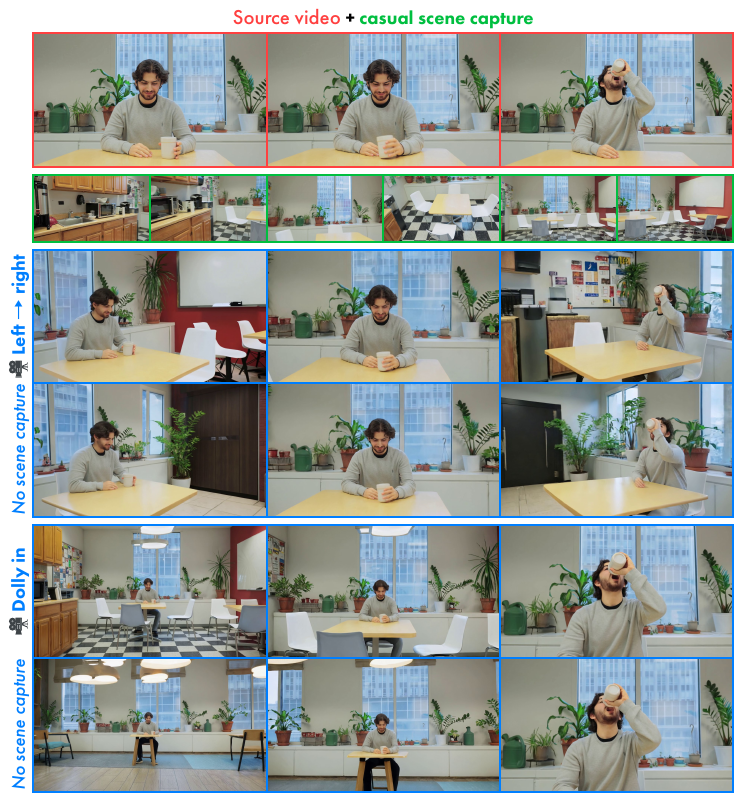}
    \caption{\Paragraphnoskip{Dynamic scene expansion} With our 4D-grounded temporally-persistent point cloud, \methodname can do video reshooting with additional scene information from casual scene captures or alternate angles by doing joint 4D reconstruction of these frames with the source video. Doing so reduces video model hallucinations and provides stronger control beyond the source video.}
    \label{fig:env_aware}
\end{figure}

\begin{figure}
    \centering
    \includegraphics[width=\columnwidth]{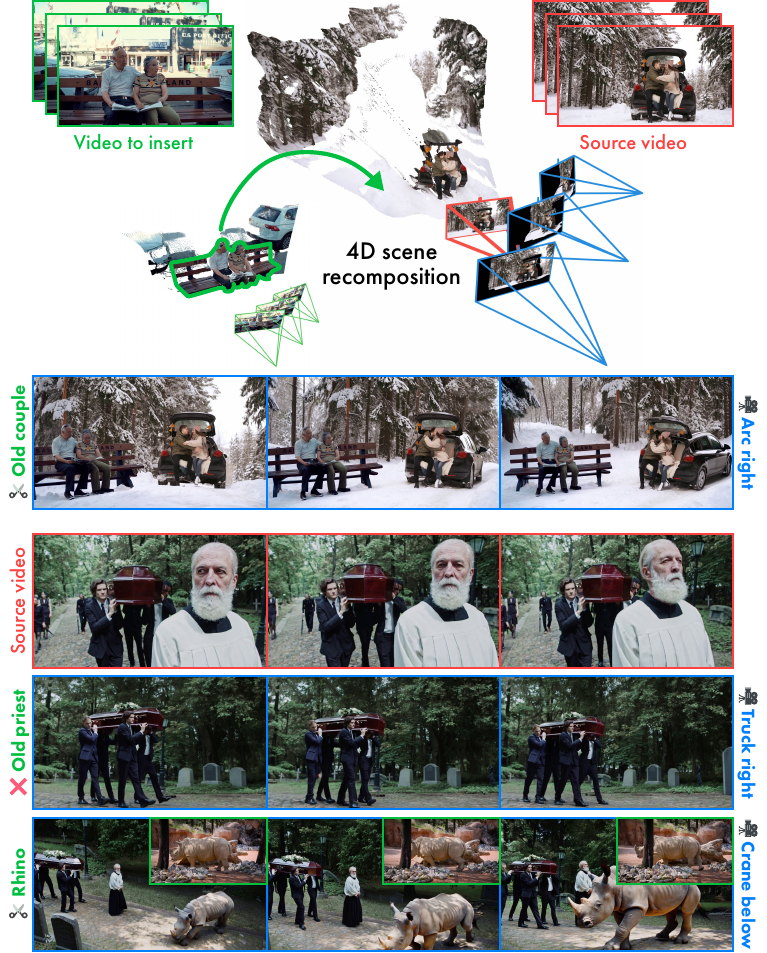}
    \caption{\Paragraphnoskip{4D scene recomposition} By directly editing the 4D point cloud, \methodname can recompose 4D scenes from the source video or other inserted videos. Importantly, our method synthesizes physically plausible lighting when inserting a rhino lit by sunlight through leaves into an otherwise overcast scene.}
    \label{fig:scene_recomp}
\end{figure}

\begin{figure*}
    \centering
    \includegraphics[width=\textwidth]{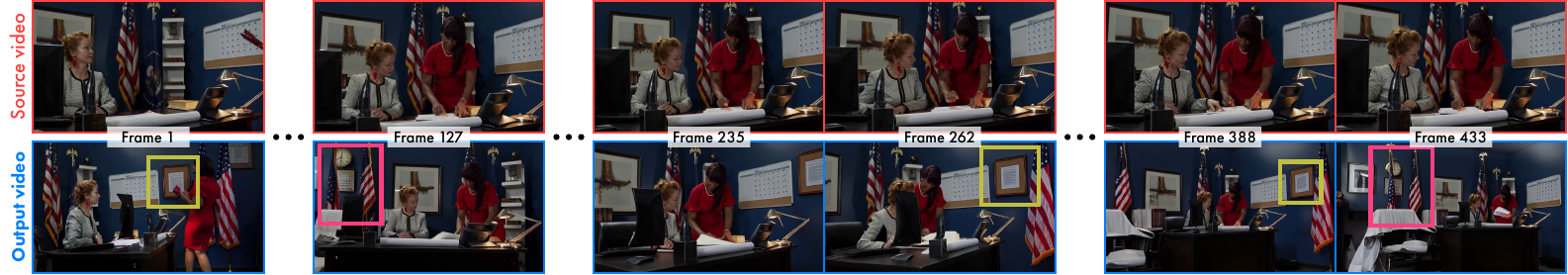}
    \caption{\Paragraphnoskip{Long video inference with memory} \methodname can reshoot long videos by doing inference in chunks. By registering static pixels of newly generated chunks back into the temporally-persistent 4D point cloud, \methodname maintains an explicit, 4D-grounded memory of generated content. We showcase this above as the camera arcs around the scene, indicated with color-matched yellow and pink boxes.}
    \label{fig:long_video}
\end{figure*}

\subsection{Applications}
\label{sec:results:app}

\Paragraph{Dynamic scene expansion}
Video reshooting requires the video diffusion models to hallucinate pixels not existent in the source video, even though we often have more visual information of the environment. For example, we may have casual captures of a scene or alternate camera angles on a film set. \methodname's explicit context grounding with temporally-persistent 4D point clouds enables us to incorporate this information by doing joint 4D reconstruction of the source video and additional scene frames. Figure \ref{fig:env_aware} shows an example of dynamic scene expansion, where the addition of temporally-persistent casual scene captures enables more faithful environment reproduction.

\Paragraph{4D scene recomposition}
As \methodname is trained to be robust to point cloud artifacts, we can directly edit and recompose the 4D point cloud to manipulate, duplicate, delete, and even insert new subjects while maintaining their dynamics. Figure \ref{fig:scene_recomp} shows examples of 4D scene recomposition. Notably, the Figure includes an example where we insert the point cloud of a rhino illuminated by sunlight through leaves into an overcast funeral procession scene. Our method naturally blends these differing lighting conditions, generating a region of dappled light around the rhino while keeping the procession in soft shadows under the trees.

\Paragraph{Long-video inference with memory}
For video reshooting on long videos beyond the video diffusion model's trained context window, our temporally-persistent 4D point cloud acts as an explicit, compressed context to retain generated static content across camera viewpoint changes. To do so, we autoregressively generate chunks of the video in target cameras that fit within our model's context window, where we train a variant of our model based on the first-frame-conditioned \texttt{Wan2.1-I2V-14B} to ensure visual consistency between chunks. Figure \ref{fig:long_video} shows an example of long-video inference that retains memory of generated content.

\section{Conclusion}
\label{sec:conclusion}

We have presented \methodname, a video reshooting framework that synthesizes the dynamic scene given by an input source video from novel camera trajectories and viewpoints. By explicitly preserving seen content from the source video with a temporally-persistent 4D point cloud and training a video diffusion model with 4D-reconstructed dynamic multiview data, our method is robust to real-world point cloud artifacts under a wide variety of input videos and target cameras. Extensive quantitative evaluations, qualitative comparisons, and the user study validate our method's 4D consistency, camera control accuracy, and video fidelity compared to state-of-the-art baselines. \methodname also generalizes to real-world applications such as dynamic scene expansion, 4D scene recomposition, and long video inference with memory.

\Paragraph{Limitations}
Though \methodname is robust to a wide variety of real-world videos and target cameras with 4D reconstruction point cloud artifacts, it lacks user control over how closely to follow a potentially imperfect point cloud as opposed to utilizing its video model prior to correct geometry. A promising extension for our work is to add a control mechanism which `interpolates' between the explicit prior (point cloud) and implicit prior (source video and camera embedding) that users can decide based on their use case.

\Paragraph{Broader impacts}
As a method building off of a video diffusion model, \methodname inherits the ethical questions which come with large generative models. Enabling camera control over any video can have a profound effect on the emotional impact and public perception of the video, raising ethical concerns about content ownership and transformative work despite the creative possibilities our method opens.

\Paragraph{Acknowledgements}
We would like to thank Aleksander Ho\l y\'nski, Wenqi Xian, Dan Zheng, Mohsen Mousavi, Li Ma, and Lingxiao Li for their technical discussions; Ryan Tabrizi, Tianyi Lorena Yan, and Shreyas Havaldar for appearing in our demo videos;  Lukas Lepicovsky, David Rhodes, Nhat Phong Tran, Dacklin Young, and Johnson Thomasson for their production support; Jeffrey Shapiro, Ritwik Kumar, and Hossein Taghavi for their executive support; Jennifer Lao and Lianette Alnaber for their operational support.

{
    \small
    \bibliographystyle{unsrtnat}
    \bibliography{main}
}

\clearpage
\maketitlesupplementary
\appendix

\section{More qualitative results on video reshooting} \label{supp:qualitative}

We show more qualitative results in this section. We recommend viewing the results as videos on our \projectpage, which also contains more results than the paper does.

\begin{figure*}
    \centering
    \includegraphics[width=\textwidth]{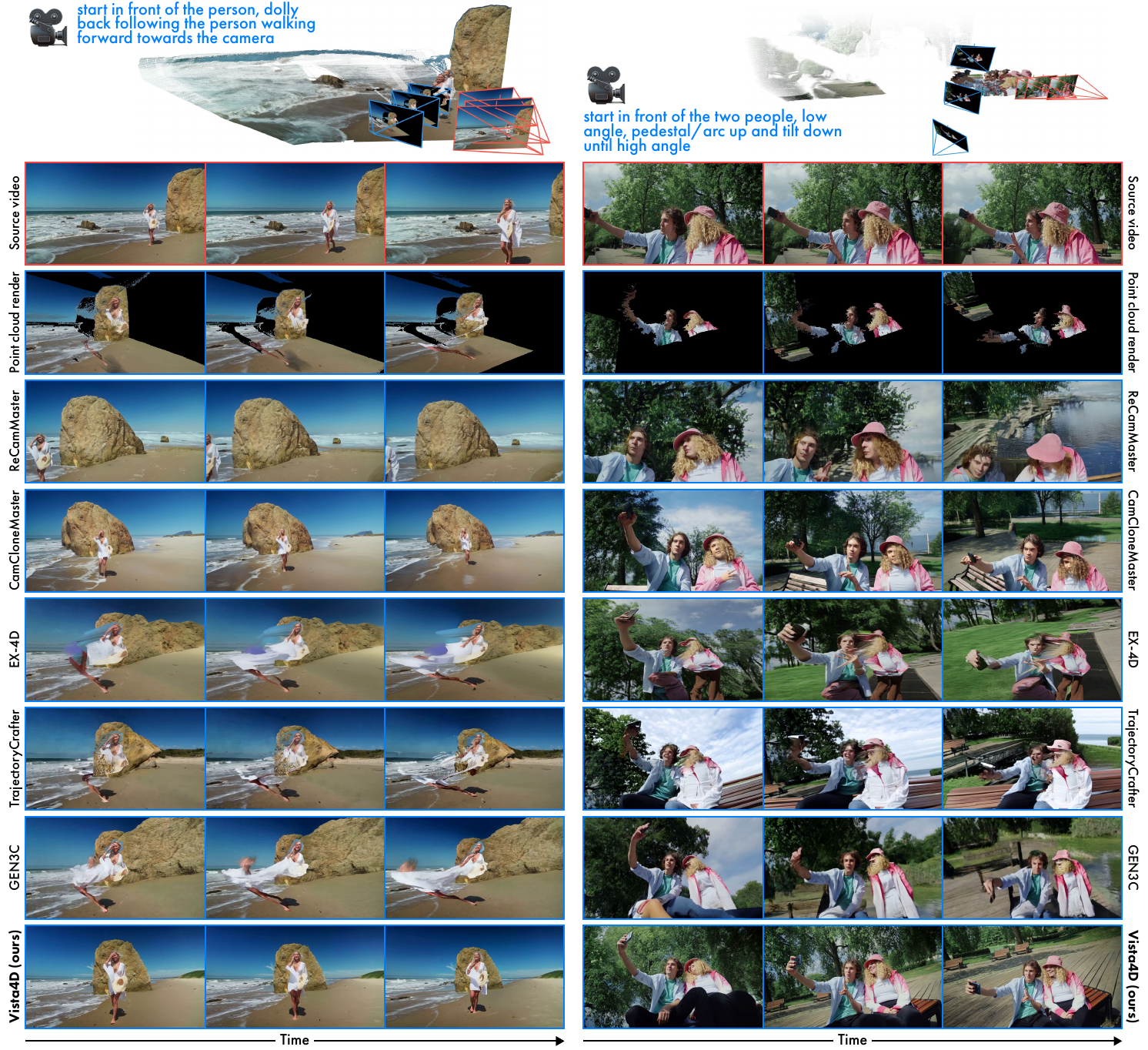}
    \caption{\Paragraphnoskip{More qualitative comparison on real-life monocular videos, part 1/2} We show two more video reshooting examples of \methodname compared to baselines: ReCamMaster \cite{bai2025recammaster}, CamCloneMaster \cite{luo2025camclonemaster}, EX-4D \cite{hu2025ex4d}, TrajectoryCrafter \cite{yu2025trajcraft}, and GEN3C \cite{ren2025gen3c}. We encourage viewing these comparisons as videos on our \projectpage, which also contains more comparisons.}
    \label{fig:qual_comp_1}
\end{figure*}

\begin{figure*}
    \centering
    \includegraphics[width=\textwidth]{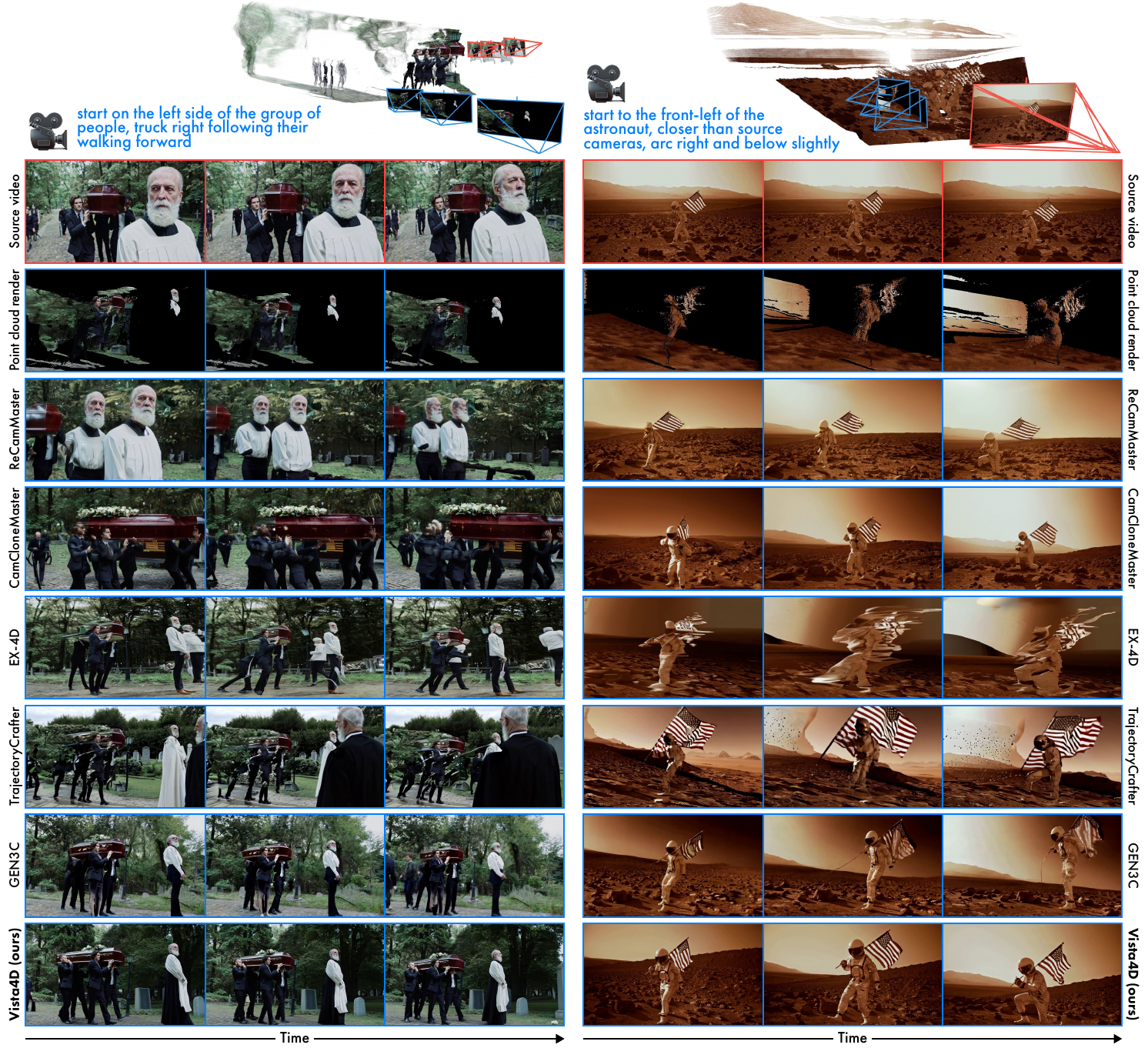}
    \caption{\Paragraphnoskip{More qualitative comparison on real-life monocular videos, part 2/2} We show two more video reshooting examples of \methodname compared to baselines: ReCamMaster \cite{bai2025recammaster}, CamCloneMaster \cite{luo2025camclonemaster}, EX-4D \cite{hu2025ex4d}, TrajectoryCrafter \cite{yu2025trajcraft}, and GEN3C \cite{ren2025gen3c}. We encourage viewing these comparisons as videos on our \projectpage, which also contains more comparisons.}
    \label{fig:qual_comp_2}
\end{figure*}

\Paragraph{Comparison to baselines}
We show more qualitative comparisons of \methodname to baselines ReCamMaster \cite{bai2025recammaster}, CamCloneMaster \cite{luo2025camclonemaster}, EX-4D \cite{hu2025ex4d}, TrajectoryCrafter \cite{yu2025trajcraft}, and GEN3C \cite{ren2025gen3c}, in Figures \ref{fig:qual_comp_1} and \ref{fig:qual_comp_2}. \methodname consistently has better preservation of the source video, more accurate camera control, and better video fidelity. Even more comparisons to baselines can be found as videos on our \projectpage.

\begin{figure*}
    \centering
    \includegraphics[width=\textwidth]{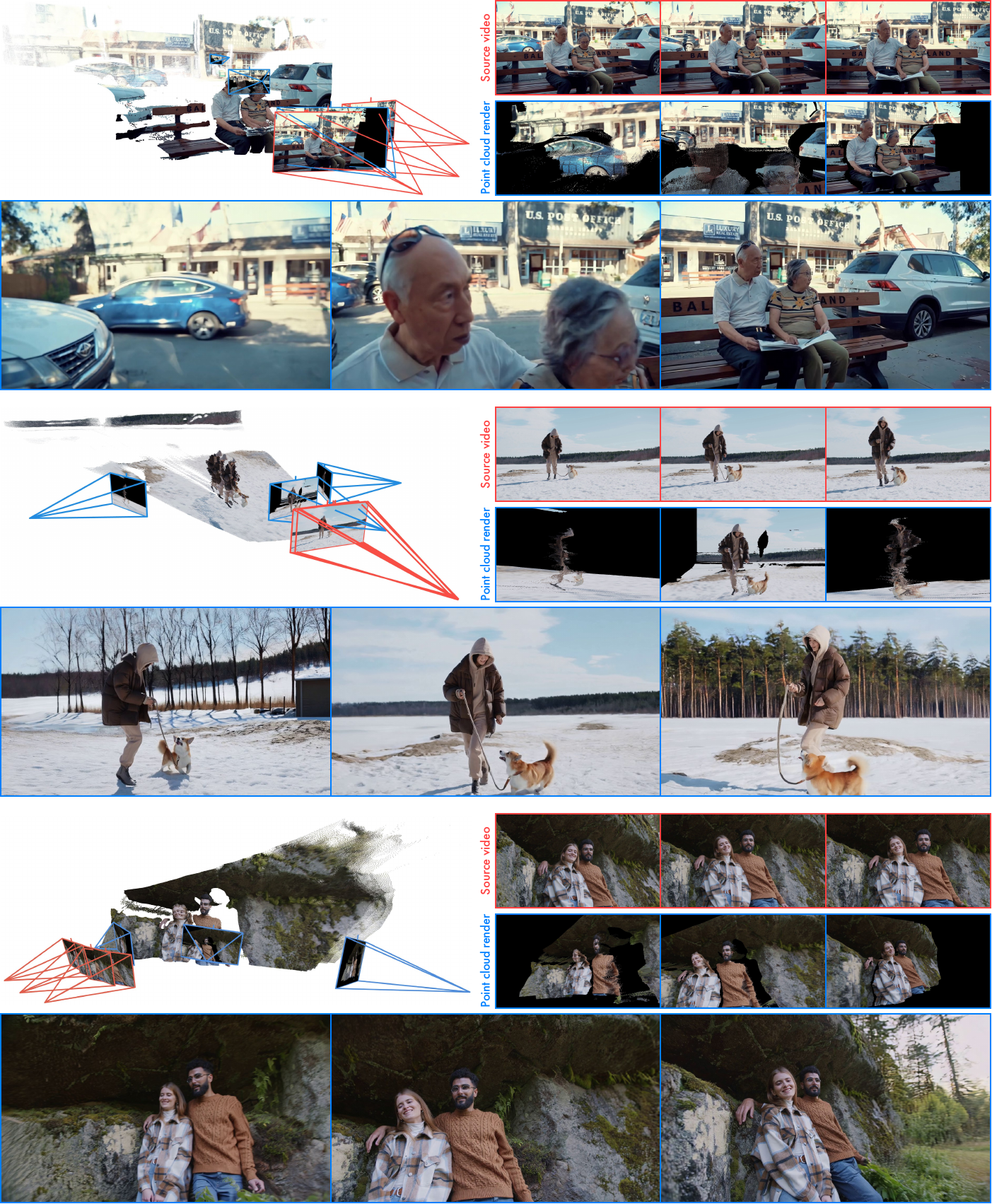}
    \caption{\Paragraphnoskip{Video reshooting results at 720p} We show video reshooting results of \methodname with our $1280 \times 720$ finetuned checkpoint. More 720p results of our method can be found as videos on our \projectpage.}
    \label{fig:results_720p}
\end{figure*}

\Paragraph{Video reshooting at 720p} Figure \ref{fig:results_720p} shows video reshooting results of \methodname with our $1280 \times 720$ finetuned checkpoint. More 720p video reshooting results can be found on our \projectpage.

\subsection{More application results and details}

\begin{figure*}
    \centering
    \includegraphics[width=\textwidth]{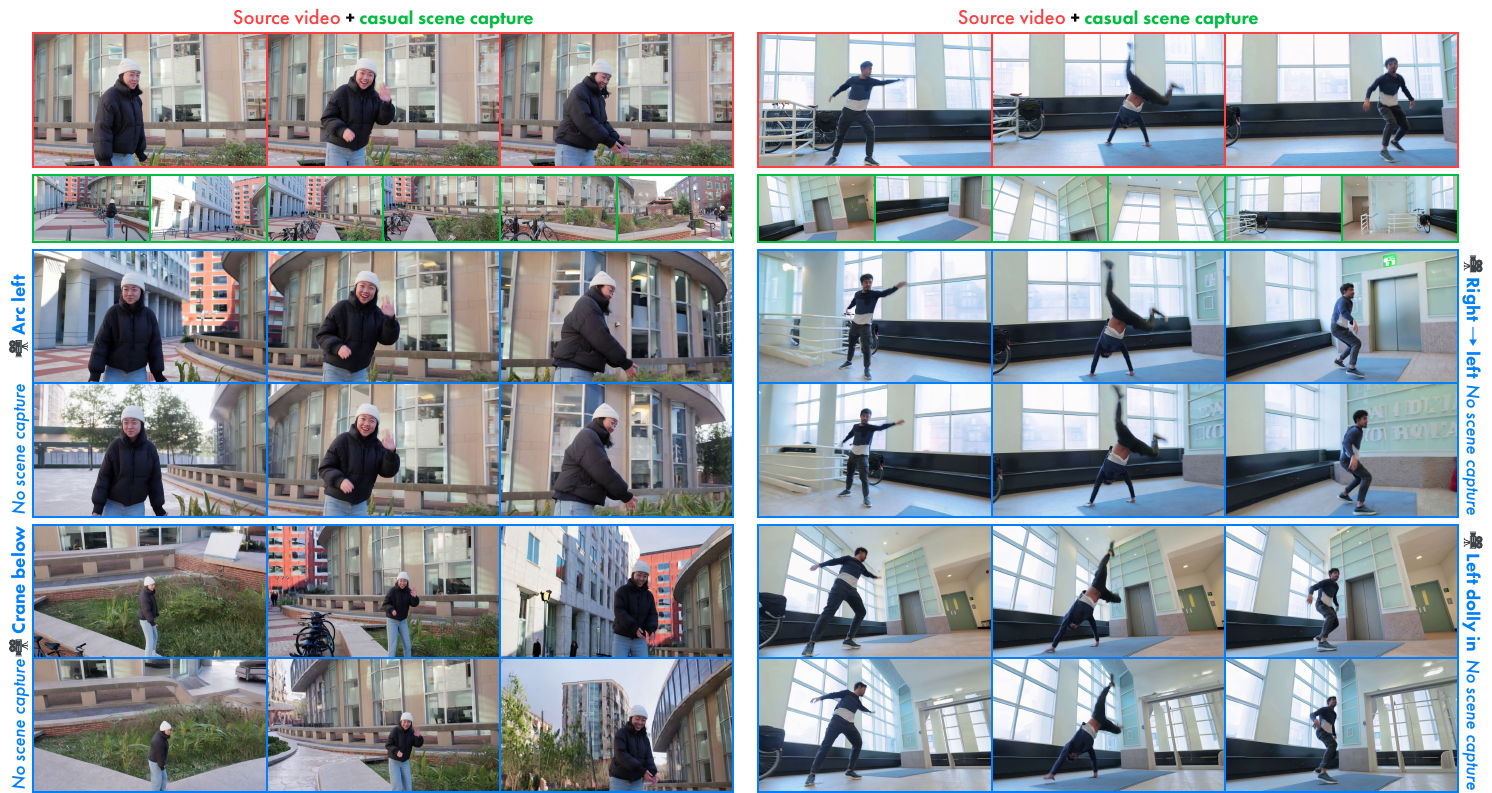}
    \caption{\Paragraphnoskip{More dynamic scene expansion results} We show more dynamic scene expansion results, where we incorporate additional scene information from casual scene captures by doing joint 4D reconstruction of these frames with the source video. We encourage viewing these results (and more) as videos on our \projectpage.}
    \label{fig:multiview_1}
\end{figure*}

\Paragraph{Dynamic scene expansion}
We show more dynamic scene expansion results in Figure \ref{fig:multiview_1}, where we incorporate additional scene information from casual scene captures by doing joint 4D reconstruction of these frames with the source video. We encourage viewing these results (and more) as videos on our \projectpage.

\begin{figure*}
    \centering
    \includegraphics[width=\textwidth]{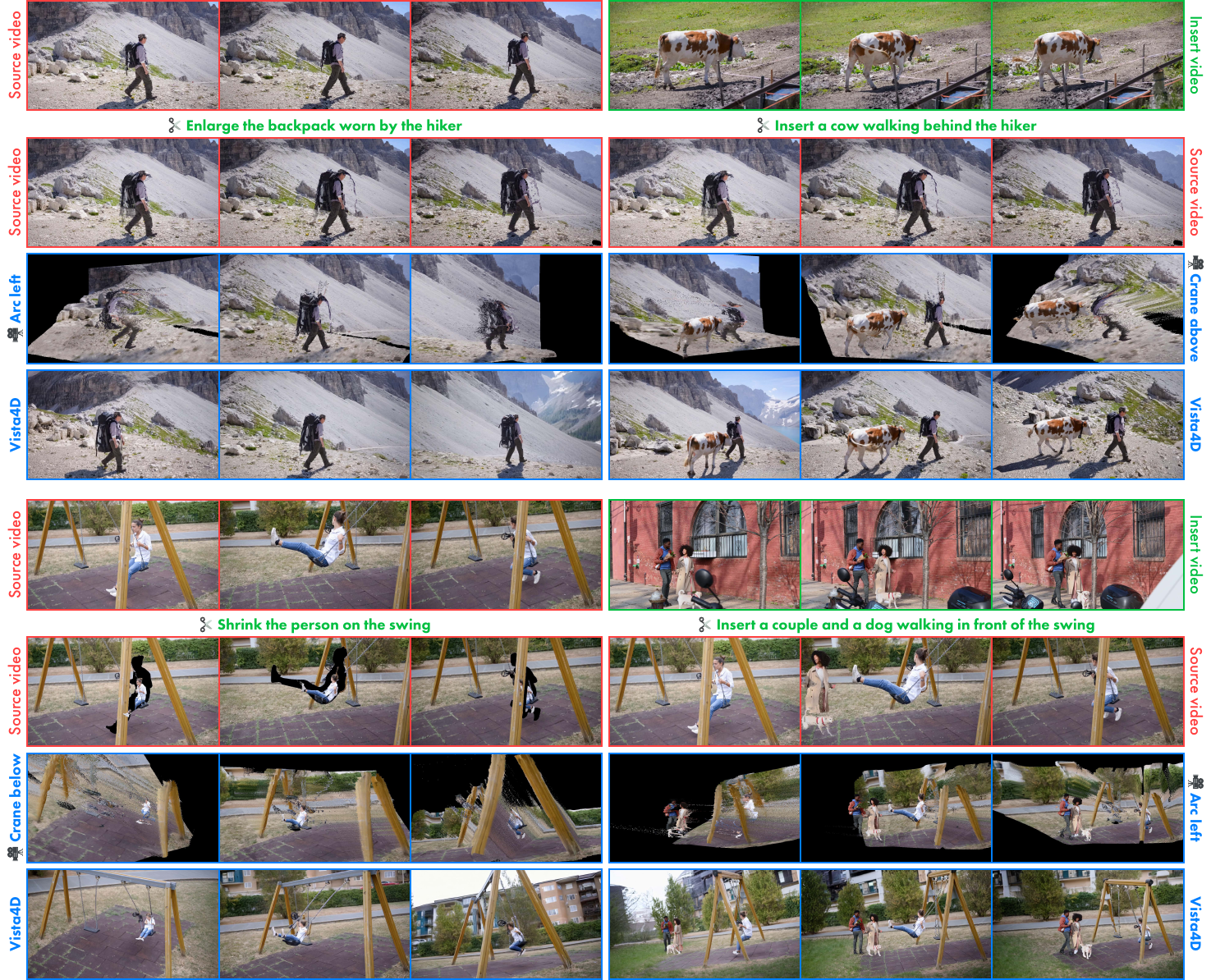}
    \caption{\Paragraphnoskip{More 4D scene recomposition results} We show more 4D scene recomposition results by directly manipulating the 4D point cloud. To prevent conditioning conflicts between the \emph{unedited} source video and render of the \emph{edited} point cloud, we instead condition on the \emph{edited} source video which is just the edited point cloud rendered from the source cameras. We encourage viewing these results (and more) as videos on our \projectpage.}
    \label{fig:scene_recomp_1}
\end{figure*}

\Paragraph{4D scene recomposition}
We show more 4D scene recomposition results in Figure \ref{fig:scene_recomp_1}. To prevent conditioning conflicts between the \emph{unedited} source video and render of the \emph{edited} point cloud, we instead condition on the \emph{edited} source video which is just the edited point cloud rendered from the source cameras without static pixel temporal persistence. Since, during training, we still include the source video condition for monocular videos, where the source video is the first render $\mathbf{X}^\tgtsrc$ of double reprojection, our model is also robust to holes and slight artifacts in the source video which the \emph{edited} source videos can contain. We encourage viewing these results (and more) as videos on our \projectpage.

\begin{figure*}
    \centering
    \includegraphics[width=\textwidth]{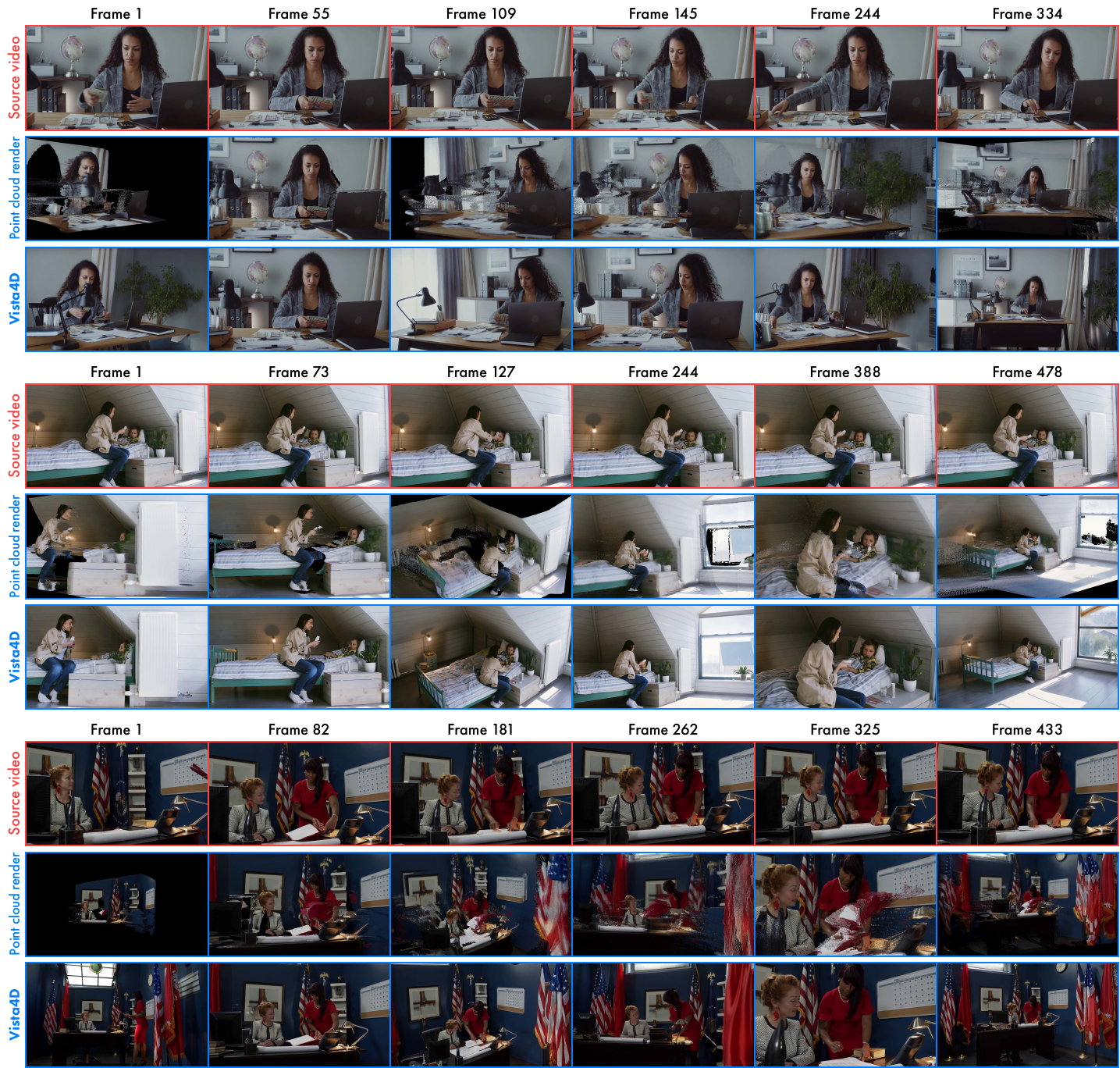}
    \caption{\Paragraphnoskip{More long video inference results} We show more results of inference on long videos. To do so, we chunk the source video into $49$-frame clips and run inference clip-by-clip. To explicitly preserve generated content, we continuously integrate the point cloud from the newly generated video into the existing one after each inference pass, which is visualized in the point cloud renders above. We encourage viewing these results (and more) as videos on our \projectpage.}
    \label{fig:long_video_1}
\end{figure*}

\Paragraph{Long video inference}
We show more long video inference results in Figure \ref{fig:long_video_1} and on our \projectpage. To support \methodname inference on long source videos, we chunk the source video into $49$-frame clips and run inference clip-by-clip. To hold explicit memory of generated content across clips, we need a 4D reconstruction method to continuously integrate the dynamic point cloud from the newly generated video clip into the existing one after each inference pass. We find existing autoregressive 4D reconstruction methods~\cite{wang2025cut3r, lan2025stream3r} to suffer from bad accuracy with very long videos, while the more accurate $\pi^3$~\cite{wang2025pi3} model only supports feedforward reconstruction. Therefore, we extend $\pi^3$ to support chunk-autoregressive inference by subsampling a constant number of frames from the existing frames and concatenating them with the newly generated frames for joint reconstruction. We then fit the existing-frame part of the predicted cameras to the known camera parameters using Umeyama alignment \cite{umeyama2002least}, while registering the new camera poses and point clouds from the generated frames to the 4D reconstruction result.

\begin{figure*}
    \centering
    \includegraphics[width=0.75\textwidth]{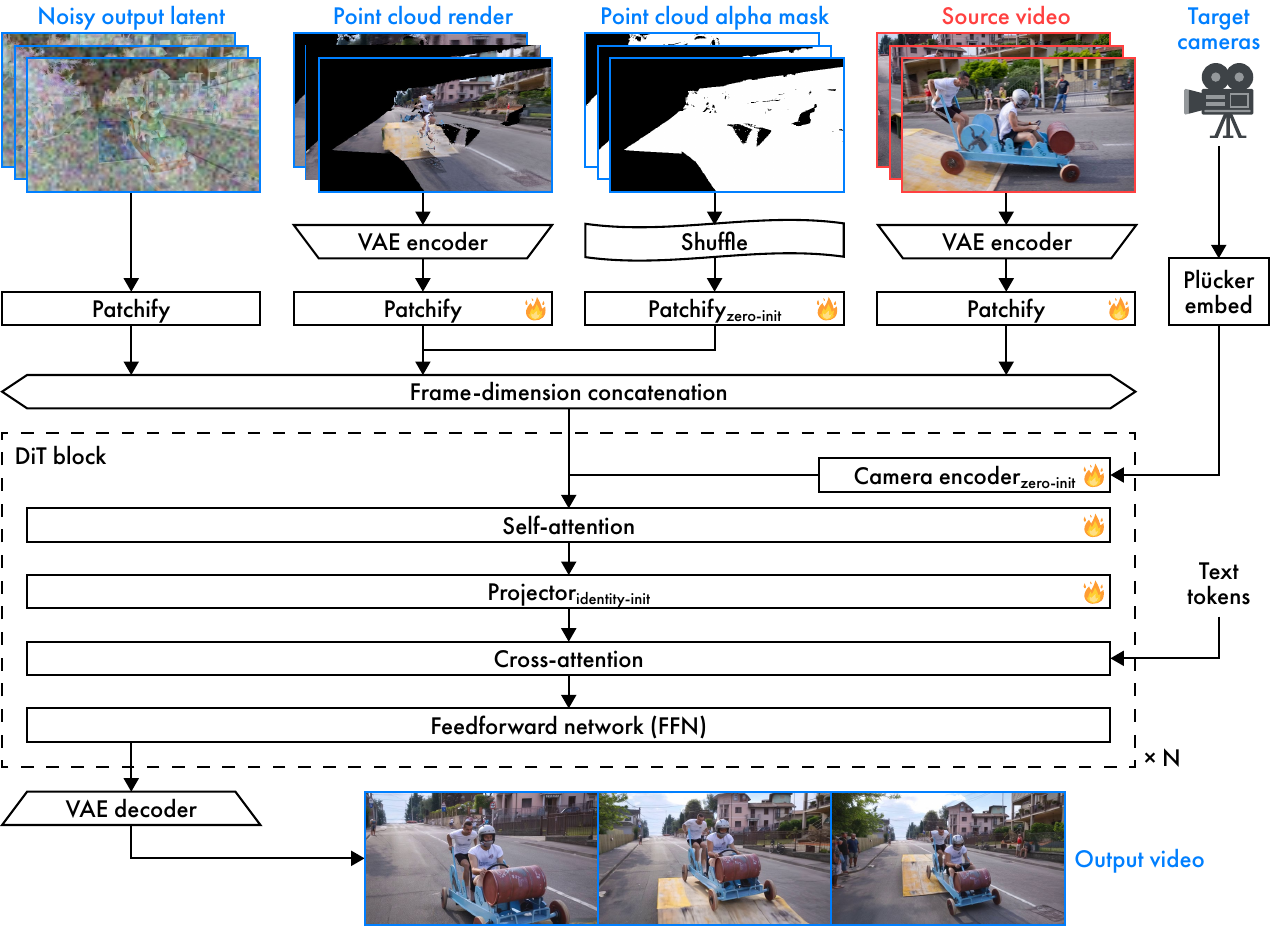}
    \caption{\Paragraphnoskip{Model architecture} The above diagram shows the model architecture for \methodname. The fire icon indicates trainable parameters. We build upon \texttt{Wan2.1-T2V-14B} \cite{wan2025}, and we omit timestep conditioning, text prompt to token embedding, modulation, layer normalization, output unshuffle, and diffusion model denoising in the diagram for simplicity. All patchify layers are initialized from the base video model besides that of the point cloud render alpha mask, which is zero-initialized. The camera encoder is zero-initialized, and the projector after self-attention is initialized as the identity affine transformation.}
    \label{fig:architecture}
\end{figure*}

\section{Model architecture details} \label{supp:architecture}

The model architecture for \methodname builds upon a text-to-video (T2V) diffusion model, namely \texttt{Wan2.1-T2V-14B} \cite{wan2025}. We finetune the T2V model to be additionally conditioned on an input source video, target cameras, point cloud render in said target cameras, and the point cloud render's alpha mask, where the model produces the output video which synthesizes the dynamic scene represented by the source video from the given target cameras. A diagram of our model architecture can be found in Figure \ref{fig:architecture}.

We first encode the source video and point cloud render into latents with the VAE, while token shuffling the point cloud render's alpha mask to match the latent space height and width. We initialize patchify layers for the source video and point cloud render from the base video model, and we zero-initialize the alpha mask's patchify layer to sum the resulting tokens with that of the point cloud render. We then concatenate the source video and point cloud render tokens with that of the noisy target along the frame dimension.

For each DiT block, we inject the target cameras as Pl\"{u}cker embeddings \cite{kuang2024collaborative, he2025cameractrl, xu2025virtuallybeing} via a zero-initialized linear camera projection which is summed with the hidden states before self-attention. After self-attention, we additionally project the hidden states with an identity-initialized affine transform before cross-attention and feedforward network (FFN), inspired by ReCamMaster \cite{bai2025recammaster}. We train the camera encoders, self-attention, projector, and all patchify layers besides that of the noisy output latent, while freezing all other parameters.

\section{Dataset and training details}
\label{sec:supp_train_data}

\subsection{Training dataset}

We train with a combination of multiview and monocular video datasets. For multiview videos, MultiCamVideo \cite{bai2025recammaster} is a synthetic time-synchronized dynamic multiview dataset from ReCamMaster. For monocular videos, \texttt{OpenVidHD-0.4M} \cite{nan2025openvid} is a filtered and labeled monocular video dataset of internet videos. The sampling ratio of multiview and monocular videos is $1:1$. More details for how we process each dataset are below.

\Paragraph{MultiCamVideo} MultiCamVideo renders its scenes at fixed camera intrinsics, and the dataset contains four unique intrinsics. For each of these intrinsics, we select the first $512$ (out of $3400$) scenes. We run 4D reconstruction with STream3R \cite{lan2025stream3r} with a moving window size of $128$ so all ten views of MultiCamVideo can fit on a single GPU. Then, we input the frames in a frame-first (\ie, we stream in all views for the current frame before moving onto the next frame) as opposed to a view-first (\ie, we stream in all frames for the current view before moving onto the next view) order. Formally, the frame-first order rearranges the tensor from \texttt{v f h w 3} to \texttt{(f v) h w 3}, where \texttt{v} is the view dimension. We do so as we find the frame-first order better ensures rough foreground/dynamic subject depth alignment between different views (as the relative scale of foreground subjects to the background scene is inherently ambiguous) since STream3R is processing different views from the same frame in close proximity due to the moving window. Though this results in temporal jittering of the predicted target cameras, we simply smooth the target camera intrinsics and extrinsics with a Gaussian kernel at the end as MultiCamVideo only renders with smooth cameras. We caption the videos with a combination of \texttt{cogvlm2-video-llama3-chat} and \texttt{cogvlm2-llama3-caption} \cite{hong2024cogvlm2, yang2025cogvideox}.

\Paragraph{\texttt{OpenVidHD-0.4M}} We select a random 60K subset from the dataset. As OpenVid provides high-level camera movement annotations, we filter for videos that are not labeled \texttt{"static"} to better learn more dynamic target cameras. We further filter out video cuts in the downloaded videos with \texttt{PySceneDetect} \cite{castellano2025pyscenedetect}. We use captions from the dataset.

Inspired by Uni4D \cite{yao2025uni4d}, we automatically segment dynamic pixels from our all datasets to produce the static pixel masks for constructing our temporally-persistent point clouds. For each video, we obtain semantic classes with RAM \cite{zhang2023ram} and prompt \texttt{Llama-3.1-8B-Instruct} \cite{llama2024llama3} to filter for subjects/nouns that would reasonably be dynamic in a video. With the list of keywords, we segment per-frame dynamic pixels with Grounded SAM 2 \cite{ravi2024sam2, ren2024groundingdino, ren2024groundedsam} and invert the result to obtain our static pixel masks.

\subsection{Training details}

We finetune \texttt{Wan2.1-T2V-14B} \cite{wan2025} for our main checkpoint. We do $10\%$ random drops each for the source video, point cloud render, prompt, and camera conditioning. When dropping the source video and/or the point cloud render, we set their latents as Gaussian noise following ReCamMaster \cite{bai2025recammaster} and zero their corresponding alpha masks.

\Paragraph{Removing the matching-first-frame constraint}
Most of the baselines that we compare to in this paper assume that the first frame of the source and target cameras (intrinsics and extrinsics) match. Doing so enables most of them to finetune from an image-to-video (I2V) as opposed to text-to-video (T2V) diffusion model to utilize the strong preservation and geometry priors of the first-frame-conditioned model. Additionally, for implicit-prior methods, this enables the camera conditioning to be relative to the first source video frame as opposed to some translation- and scale-invariant world space. We do not have this constraint for \methodname, which is achieved by both using a text-to-video model and also data processing. Notably, since MultiCamVideo always has matching source and target camera first frames, we do $50\%$ random time-reversal of the source and target videos together.

\Paragraph{Finetuning I2V diffusion models for long video inference}
We also finetune \texttt{Wan2.1-I2V-14B} \cite{wan2025} for long video inference, as we find the first-frame condition helpful for maintaining visual consistency between consecutive inference chunks. We train with the exact same dataset and simply also condition the model on the first frame of the target video, even if said first frame does not match that of the source video. We do $30\%$ random drop for the image condition to strengthen the point cloud render's influence, as we otherwise observe poorer camera control during preliminary experiments when the model more heavily relies on the image condition. We also apply noise-augmentation condition on the image latent to reduce quality degradation with more inference chunks. Namely, given the image condition $\mathbf{X}^\img$, we obtain the augmented $\overline{\vec{X}^\img} = (1 - \alpha) \vec{X}^\img + \alpha \boldsymbol{\epsilon}$ where $\alpha = 0.05$ and $\boldsymbol{\epsilon} \sim \NN(0, \vec{I})$. During inference, we use our T2V-finetuned checkpoint for the first $49$-frame clip and I2V-finetuned checkpoint for all subsequent clips.

\section{Evaluation dataset and user study details} \label{supp:eval_user}

We construct a $110$ video-camera pair evaluation dataset for quantitative evaluations and our user study. We select $13$ videos from DAVIS \cite{perazzi2016davis} and $38$ videos from Pexels \cite{pexels2025} which are high quality and contain dynamic scene and/or camera motion. We then design two to three target camera trajectories and zooms for each video with our camera design UI. We reconstruct all videos with $\pi^3$ \cite{wang2025pi3} and manually annotate keywords for segmenting dynamic pixels with Grounded SAM 2 \cite{ravi2024sam2, ren2024groundingdino, ren2024groundedsam}. We caption all videos with a combination of \texttt{cogvlm2-llama3-caption} \cite{yang2025cogvideox} and Gemini 2.5 Pro \cite{2025gemini25}. We release the evaluation dataset and our annotations with our public code and weights release, which can be found on our \projectpage.

\begin{figure*}
    \centering
    \includegraphics[width=0.75\textwidth]{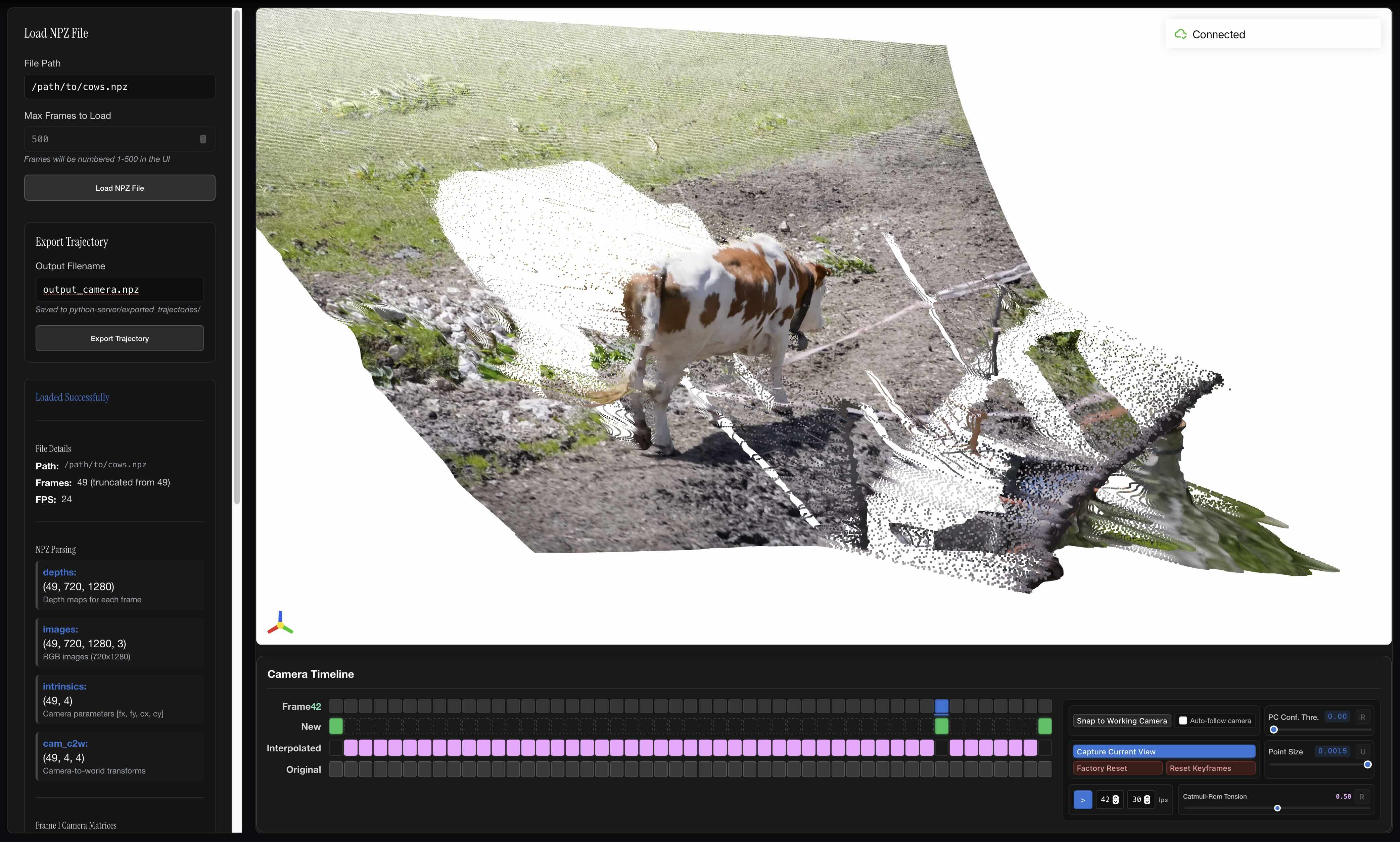}
    \caption{\Paragraphnoskip{Camera design UI} The above screenshot shows our current camera design UI, built on top of Viser \cite{yi2025viser}. Users can set camera intrinsics and extrinsics keyframes and interpolation tension/smoothness, while being able to preview the 4D reconstructed point cloud from their defined target cameras in real time.}
    \label{fig:camera_ui}
\end{figure*}

\Paragraph{Camera design UI}
We build an interface for easily defining target cameras given the 4D reconstruction of a source video, built on top of Viser \cite{yi2025viser}, and a screenshot of which can be found in Figure \ref{fig:camera_ui}. Currently, users can set camera intrinsics and extrinsics keyframes and interpolation tension/smoothness, while being able to preview their defined target cameras when playing back the video/4D point cloud. As the UI simply outputs camera intrinsics and extrinsics which are used in a separate point cloud rendering module, this does not affect temporal persistence for our final point cloud render. We release our camera design UI with our public code release, which can be found on our \projectpage.

\Paragraph{Running the baselines}
Every baseline that we compare \methodname to, besides TrajectoryCrafter \cite{yu2025trajcraft}, does not support differing first frame source and target cameras due to their data constraints or processing and/or from finetuning an I2V model (where they use the first frame of the source video as the image condition). To run inference for these baselines on video-camera pairs in our evaluation dataset which do not have matching first frame source and target cameras, we follow the following procedure first implemented in TrajectoryCrafter's codebase as \texttt{infer\_direct} mode \cite{yu2025trajcraft}: Freeze the first frame of the point cloud and move the first frame of the source camera to that of the target camera, then unfreeze the point cloud and continue the target cameras from there. In order to fairly compare with several baselines at once, we unify the quantitative evaluation at $672 \times 384$ (though we still run at each baseline's native resolutions), as several baselines do not have higher native resolutions.

\begin{table*}
    \caption{\Paragraphnoskip{Preprocessing and inference time} With user-defined dynamic keywords, inference preprocessing involves segmentation (Grounding SAM 2) and 4D reconstruction ($\pi^3$). Model inference are all 50 steps. We run everything on an NVIDIA A100 80GB.}
    \label{tab:preprocess_inference}
    \centering
    \begin{adjustbox}{max width=\linewidth}
        \begin{tabular}{l l l c c c}
            \toprule
            Method & Base model & Resolution & Segmentation (s) & 4D reconstruction (s) & Model inference (s) \\
            \midrule
            ReCamMaster & \texttt{Wan2.1-T2V-1.3B} & $832 \times 480$ & - & \multirow{2}{*}{\makecell{\textit{Implicit prior,}\\\textit{no 4D reconstruction}}} & 523.2 \\
            CamCloneMaster & \texttt{Wan2.1-T2V-1.3B} & $832 \times 480$ & - & & 1062 \\ \hline
            TrajectoryCrafter & \texttt{CogVideoX-Fun-5B} &  $672 \times 384$ & - & & 170.1 \\
            EX-4D & \texttt{Wan2.1-T2V-14B} & $672 \times 384$ & - & \multirow{4}{*}{\makecell{\textit{Explicit prior,}\\\textit{all using $\pi^3$}\\\textit{with time:}\\3.110}} & 698.9 \\
            GEN3C & \texttt{Cosmos1.0Diffusion-7BVideo2World} & $1280 \times 704$ & - & & 1110 \\
            \textbf{\methodname (ours)} & \texttt{Wan2.1-T2V-14B} & $672 \times 384$ & \multirow{2}{*}{22.75} & & 1195 \\
            \textbf{\methodname (ours)} & \texttt{Wan2.1-T2V-14B} & $1280 \times 720$ & & & 9924 \\
            \bottomrule
        \end{tabular}
    \end{adjustbox}
\end{table*}

\Paragraph{Preprocessing and inference time}
We show preprocessing and inference time of \methodname and our baselines in Table \ref{tab:preprocess_inference}, where we see that the overhead for preprocessing is negligible compared to model inference. \methodname is slower than our baselines primarily due to in-context conditioning and our slower base model. The latter may itself also contribute to higher visual quality, but as shown in our ablations in Supplementary \ref{supp:ablations}, our key designs as summarized above are the main contributors for \methodname's superior performance at specifically video reshooting.

\begin{figure*}
    \centering
    \includegraphics[width=0.49\textwidth]{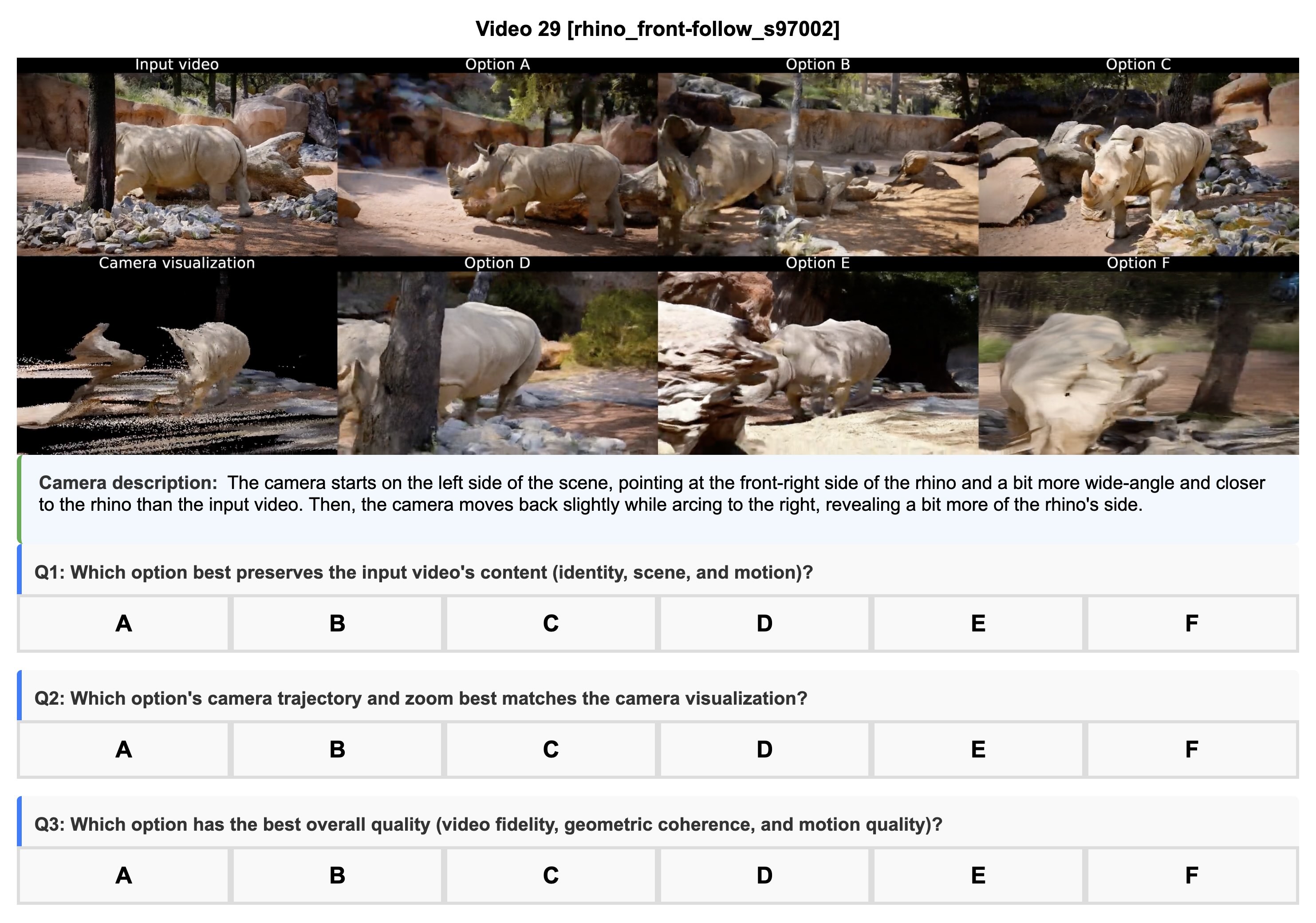}
    \includegraphics[width=0.49\textwidth]{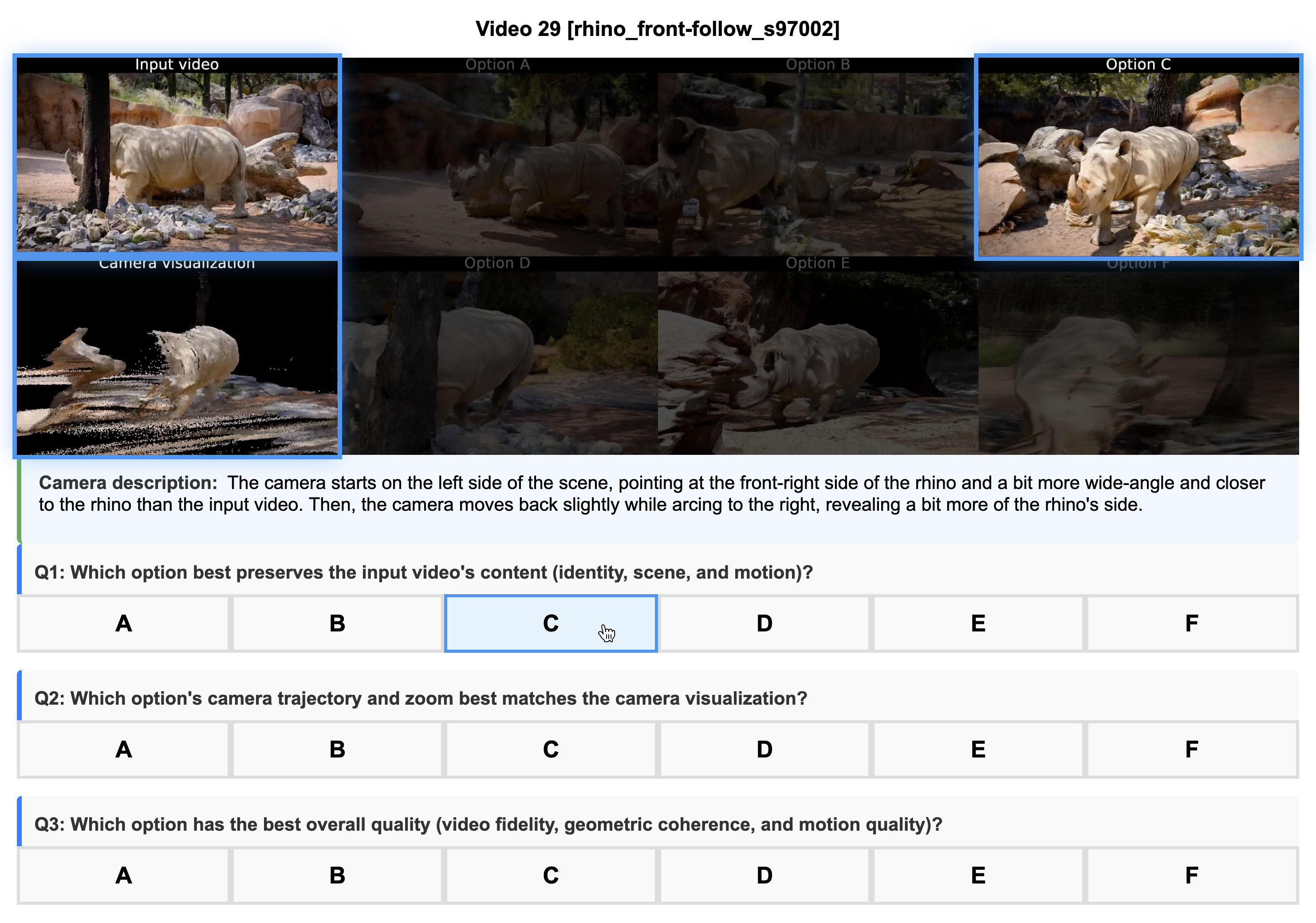}
    \caption{\Paragraphnoskip{User study} The left screenshot shows an example video-camera pair of our user study, where users are asked to select their preferred method/option on three dimensions: Source video preservation (Q1), camera control accuracy (Q2), and overall video fidelity (Q3). The right screenshot shows our user study UI highlighting the source video, point cloud render, and corresponding output video as users hover on each option. We also provide a camera description in addition to the point cloud render to communicate our intended target cameras. There are $30$ video-camera pairs in total in the user study, which was randomly selected from our $110$ video-camera pair evaluation dataset.}
    \label{fig:user_study}
\end{figure*}

\Paragraph{User study}
For our user study, we randomly select $30$ video-camera pairs from our evaluation dataset and invite $42$ participants to select their preferred method/option from \methodname and baseline video reshooting results. For each video-camera pair, we ask for the participants' preference on three dimensions: Source video content preservation (``Which option best preserves the input video's content (identity, scene, and motion)?''), camera control accuracy (``Which option's camera trajectory and zoom best matches the camera visualization?''), and overall video fidelity (``Which option has the best overall quality (video fidelity, geometric coherence, and motion quality)?''). For camera accuracy preference, we provide both the point cloud render and a short description of the intended target camera for each pair. The order of the methods is also randomized and anonymized for each pair. A screenshot of our user study can be found in Figure \ref{fig:user_study}.

\section{Quantitative evaluation metric details} \label{supp:quant_metrics}

\Paragraph{Camera control accuracy}
We perform camera control accuracy evaluation by comparing the predicted camera parameters $(\vec{R}_i^\gen, \vec{t}_i^\gen, \mathrm{FOV}_i^\gen)_{i=1}^T$ of the generated video and the target cameras $(\vec{R}_i^\tgt, \vec{t}_i^\tgt, \mathrm{FOV}_i^\tgt)_{i=1}^T$, where $\vec{R}, \vec{t}$ are the camera extrinsics and $\textrm{FOV}$ is the vertical field of view from the camera intrinsics. As the target camera parameters are represented in the source video's coordinate system, we first jointly reconstruct the camera poses of both the source and generated videos and then fit the source video part of the camera poses to the known source camera parameters using Umeyama alignment \cite{umeyama2002least}.

As many target videos generated by the baselines lack 3D consistency, traditional SFM and optimization-based methods like GLOMAP~\cite{pan2024glomap} used in prior works~\cite{CameraCtrl, bai2025recammaster} would fail to reconstruct the source and the target videos jointly. To obtain a fair camera accuracy evaluation, we adopt the learning-based 4D reconstruction method $\pi^3$~\cite{wang2025pi3} for joint reconstruction, which is also the same method used to reconstruct the source camera during inference. The evaluation metrics consist of the translation error, rotation error, and intrinsics error following~\cite{CameraCtrl, huang2025vipe} where
\begin{align}
    \text{RotErr}
    &= \frac{1}{T}\sum_{i=1}^{T}
    \arccos\!\left(
    \frac{\operatorname{tr}\!\left(\vec{R}_i^\tgt {\vec{R}_i^\gen}^{\!\top}\right) - 1}{2}
    \right) \, ,
    \\
    \text{TransErr}
    &= \frac{1}{T}\sum_{i=1}^{T}
    \left\| \vec{t}_i^\tgt - \vec{t}_i^\tgt \right\|_2^2 \, ,
    \\
    \text{IntrinsicsErr}
    &= \frac{1}{T}\sum_{i=1}^{T}
    \left\lvert \text{FOV}_i^{\text{tgt}} - \text{FOV}_i^\gen \right\rvert.
\end{align}

\Paragraph{3D consistency via reprojection error (RE@SG)} Traditional NVS metrics such as PSNR, SSIM, and LPIPS compare the generated images against a fixed-set of ground truth images. However, using them to evaluate a generative model unfairly penalizes plausible generations when they deviate from the ground truth in unseen regions, while being restrictive with evaluation datasets without ground truths. Pippo~\cite{kant2025pippo} proposed the Reprojection Error, which enables the evaluation of the 3D consistency of the generated scene from two given camera viewpoints (\ie known intrinsics and extrinsics) by utilizing LightGlue~\cite{lightglue} and SuperPoint~\cite{detone2018superpoint} for detecting 2D point matches, then triangulating them in 3D, and computing the re-projection between the 3D points and 2D correspondences. We utilize this metric to compare the 3D consistency of the baselines with that of \methodname.

\Paragraph{Novel-view synthesis}
We perform novel-view synthesis evaluation on the \texttt{iphone}~\cite{gao2022dycheck} dataset. Following TrajectorCrafter~\cite{yu2025trajcraft}, we evaluate on the five sequence subset without label errors. We use the moving camera as the source video, and the first static camera with continuous frames as the target. We also adopt the camera pose and depth map labels provided by Shape of Motion~\cite{wang2025shapeofmotion} for the point cloud input. For pixel-wise synthesis accuracy, we follow standard protocols~\cite{mildenhall2021nerf, kerbl20233d} and evaluate PSNR, SSIM, and LPIPS. As many pixels in the target video are invisible in the source video, we also compute masked metrics following Dycheck~\cite{gao2022dycheck} and evaluate mPSNR, mSSIM, and mLPIPS with covisibility masking to better compare the model's performance in preserving content in the source video. Finally, the standard novel view synthesis metrics only evaluate the frame-wise metrics without measuring the accuracy of the synthesized motion across frames. Thus, we further evaluate motion quality by comparing the ground-truth optical flow and the generated optical flow using end-point error (EPE). As ground-truth optical flow labels are not available in the \texttt{iphone} dataset, we use an off-the-shelf model, SEA-RAFT~\cite{wang2024sea}, to predict optical flow for both the ground-truth and generated videos.

\Paragraph{Video fidelity}
We evaluate the video fidelity, visual quality, and prompt alignment of \methodname\ and all baselines on our $110$ video-camera-pair dataset. For fidelity, we compute FID~\cite{heusel2017fid} and FVD~\cite{uterthiner2019fvd} between the generated videos and their corresponding source videos. We further use VBench~\cite{huang2024vbench} to assess multiple perceptual dimensions: aesthetic quality, predicted by an aesthetic model capturing frame-level layout, color richness, and visual harmony; imaging quality, which measures distortions such as over-exposure, blur, and noise; subject consistency, evaluating how stable the main subject remains across frames; background consistency, computed via CLIP feature similarity across frames to quantify temporal stability of the scene; and temporal style, which measures similarity between video features and a temporal-style description to assess motion-style coherence. In addition, we evaluate human anatomy using VBench-2.0~\cite{zheng2025vbench2}, which reports anomaly scores for the human body, hands, and face, reflecting a model’s ability to preserve anatomically consistent humans under novel camera trajectories. Finally, we use CLIP-T~\cite{radford2021clip} to measure prompt alignment.

\section{Ablation study} \label{supp:ablations}

We show samples for our ablations on no depth artifacts \& source video conditioning and no temporal persistence. Note that all ablations samples, including ones from our full method, are from checkpoints with fewer training steps than the final $672 \times 384$ checkpoint. We encourage viewing the ablation results on our \projectpage file, as it is difficult to show artifacts like temporal jittering and at times inaccurate camera control through still frames in the paper.

\begin{figure*}
    \centering
    \includegraphics[width=\textwidth]{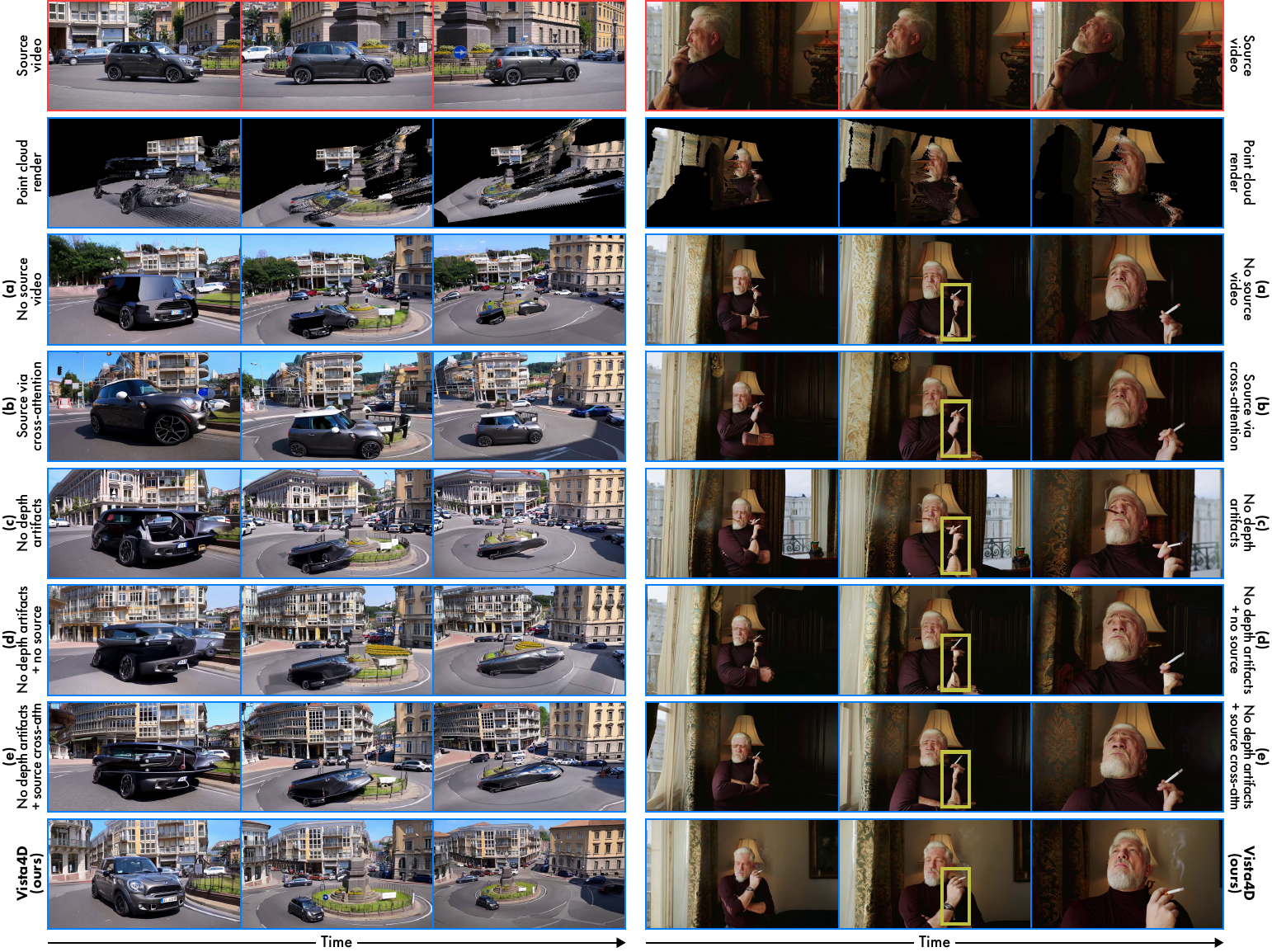}
    \caption{\Paragraphnoskip{Ablation on depth artifacts and source video conditioning} We show ablation samples on training with depth artifacts (we simulate training without depth artifacts by always doing double reprojection for point cloud rendering \cite{yu2025trajcraft}) and source video conditioning (comparing our in-context/frame-concatenated source video conditioning with no source video and source video injected via cross-attention). Both examples above show 4D reconstruction artifacts carrying over to all ablations, such as on the car (left) or the man's arm and hand (right, highlighted by yellow boxes). Notably, though injecting the source video via cross-attention can at times correct point cloud artifacts, we find that cross-attention is often not adaptive enough, such as left (b) where the car is abnormally large despite the flying back camera. Both training without depth artifacts and in-context-conditioned source video also result in temporal jittering, but that is difficult to show as still frames in the paper. Thus, we encourage viewing these (and more) ablation samples as videos on our \projectpage.}
    \label{fig:ab_das}
\end{figure*}

\subsection{Depth artifacts and source video conditioning}

We show samples of our method with ablations in Figure \ref{fig:ab_das} on the following design choices:
\begin{enumerate}[label={\alph*)}]
    \item \textbf{No source video:} We take away source video conditioning and only condition on the point cloud render.
    \item \textbf{Source video via cross-attention:} Instead of in-context conditioning (\ie, self-attention through frame concatenation), we condition the source video via cross-attention following TrajectoryCrafter \cite{yu2025trajcraft}, as it is our only explicit-prior baseline which conditions on source videos in addition to point cloud renders.
    \item \textbf{No depth artifacts:} For the multiview dynamic dataset (MultiCamVideo dataset from ReCamMaster), instead of the point cloud render being rendered the source video in the cameras of the target video, we always do double reprojection to remove depth artifacts so the point cloud render is always spatially aligned with the target video.
    \item \textbf{No depth artifacts + no source video:} Combination of (b) and (a).
    \item \textbf{No depth artifacts + source cross-attn:} Combination of (c) and (a).
\end{enumerate}

We observe two major artifacts/problems when we remove depth artifacts and/or in-context/self-attention source video conditioning during training:
\begin{enumerate}
    \item \textbf{Geometry artifacts from imprecise depth estimation:} The model is unable to correct obvious depth estimation artifacts and thus produce output artifacts.
    \item \textbf{Temporal jittering:} One artifact of real-world depth estimation/4D reconstruction is temporal jittering of the resulting point cloud. Here, the model is unable to correct this jittering. Note that this is difficult to show as still frames in the paper, so we encourage viewing the ablation samples on our \projectpage.
\end{enumerate}
Figure \ref{fig:ab_das}, left exemplifies observation 1, where the 4D reconstruction artifacts on the car carried over to all ablations, except (b) where, though there are little depth artifacts, cross-attention was unable to properly transfer the car's geometry from the source video while ensuring its correct size with the camera flying back. Figure \ref{fig:ab_das}, right also shows observation 1, where every ablation (a) to (e) displays artifacts from depth estimation, especially around the man's right arm and hand. Observation 2 (temporal jittering) is difficult to present as still frames in the paper, so we encourage viewing the ablation samples, along with more ablation results, as videos on our \projectpage.

\begin{figure*}
    \centering
    \includegraphics[width=\textwidth]{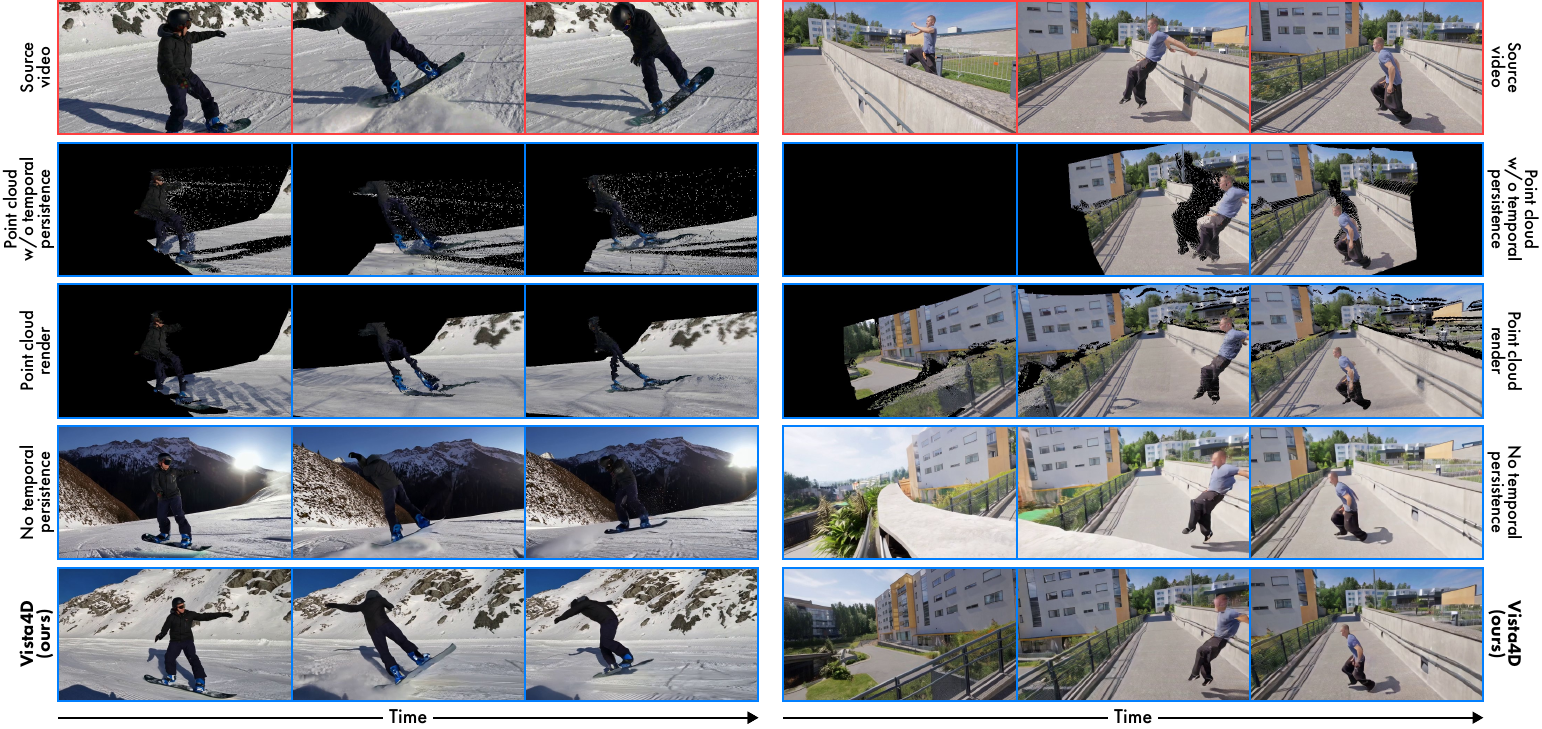}
    \caption{\Paragraphnoskip{Ablation on static pixel temporal persistence} We show ablation samples on training with and without static pixel temporal persistence. Both examples above show the no-temporal-persistence model struggling to preserve seen content from the source video, such as the snow and rock mountain (left) and the metal fence and the road beyond it (right). The model without temporal persistence also exhibits inaccurate camera control for both samples above, which is difficult to show as still frames in the paper. Thus, we encourage viewing these (and more) ablation samples as videos on our \projectpage.}
    \label{fig:ab_tp}
\end{figure*}

\subsection{Temporal persistence}

We show samples of our method trained with and without point cloud static pixel temporal persistence in Figure \ref{fig:ab_tp}, and we also show the corresponding point cloud conditioning with or without temporal persistence. We observe two major artifacts/problems when we remove temporal persistence:
\begin{enumerate}
    \item \textbf{Not preserving seen (static) content:} The model struggles to preserve seen content from the source video.
    \item \textbf{Imprecise camera control:} The model has less accurate camera control during target camera frames which have little overlap with the source video point cloud. Note that this can be difficult to show as still frames in the paper, so we encourage viewing the ablation samples on our \projectpage.
\end{enumerate}
Figure \ref{fig:ab_tp}, left encompasses observation 1, where the no-temporal-persistence model struggles to faithfully synthesize the snow and rock mountain behind the snowboarder as the per-frame point cloud render never explicitly sees it. Figure \ref{fig:ab_tp}, right also showcases observation 1, where the no-temporal-persistence model struggles to preserve the right side of the scene. Both examples also display observation 2, \ie, imprecise camera control, but that is difficult to show as still frames in the paper. Thus, we encourage viewing the ablation samples, along with more ablation results, as videos on our \projectpage.

\end{document}